%% file: main.tex
\pdfoutput=1

\documentclass[11pt]{article}

\usepackage[final]{acl}

\usepackage{times}
\usepackage{latexsym}
\usepackage{multicol, multirow}
\usepackage{booktabs}

\usepackage[T1]{fontenc}

\usepackage[utf8]{inputenc}

\usepackage{microtype}

\usepackage{inconsolata}
\usepackage{arydshln}
\usepackage{soul}

\usepackage{graphicx}
\usepackage{subcaption}
\usepackage{listings}
\usepackage{pifont}

\definecolor{codegreen}{rgb}{0,0.6,0}
\definecolor{codegray}{rgb}{0.5,0.5,0.5}
\definecolor{codepurple}{rgb}{0.58,0,0.82}
\definecolor{backcolour}{rgb}{0.95,0.95,0.92}
\definecolor{mylightblue}{rgb}{0.68, 0.85, 0.9}
\definecolor{mywheat}{rgb}{0.96, 0.87, 0.7}

\lstdefinestyle{pythonstyle}{
    backgroundcolor=\color{backcolour},   
    commentstyle=\color{codegreen},
    keywordstyle=\color{magenta},
    numberstyle=\tiny\color{codegray},
    stringstyle=\color{codepurple},
    basicstyle=\ttfamily\footnotesize\tiny,
    breakatwhitespace=false,         
    breaklines=true,                 
    captionpos=b,                    
    keepspaces=true,                 
    numbers=left,                    
    numbersep=5pt,                  
    showspaces=false,                
    showstringspaces=false,
    showtabs=false,                  
    tabsize=2
}

\lstdefinestyle{prompt_style}{
    frame=single,
    basicstyle=\ttfamily\scriptsize,
    backgroundcolor=\color{white},
    stringstyle=\color{black},
    commentstyle=\color{darkgreen}\slshape,
    stringstyle=\color{darkred},
    numberstyle=\tiny\color{codegray},
    emphstyle=\color{pink}\underbar,
    breakindent=0pt,
    escapeinside={(*@}{@*)},
    breakatwhitespace=true,
    breaklines=true,
    captionpos=b,
    keepspaces=true,
    numbersep=5pt,
    showspaces=false,                
    showstringspaces=false,
    showtabs=false,
    tabsize=2,
}

\usepackage{enumitem}
\setlist[itemize]{leftmargin=10pt}
\newcommand{\minisection}[1]{\vspace{1pt}\noindent\textbf{#1}}
\input{math_commands}

\input{commands}

\title{\textsc{RuleArena}: A Benchmark for Rule-Guided Reasoning with LLMs in Real-World Scenarios}

\author{
 \textbf{Ruiwen Zhou\textsuperscript{1}},\;
 \textbf{Wenyue Hua\textsuperscript{1}}, \;
 \textbf{Liangming Pan\textsuperscript{2}}, \;
 \textbf{Sitao Cheng\textsuperscript{1}},
\\
 \textbf{Xiaobao Wu\textsuperscript{1,3},} \;
 \textbf{En Yu\textsuperscript{1},} \;
 \textbf{William Yang Wang\textsuperscript{1}}
\\
 \textsuperscript{1}University of California, Santa Barbara \;\;
 \textsuperscript{2}University of Arizona
 \\
 \textsuperscript{3}Nanyang Technological University
}

\begin{document}
\maketitle
\begin{abstract}
This paper introduces \textsc{RuleArena}, a novel and challenging benchmark designed to evaluate the ability of large language models (LLMs) to follow complex, real-world rules in reasoning. Covering three practical domains---airline baggage fees, NBA transactions, and tax regulations---\textsc{RuleArena} assesses LLMs' proficiency in handling intricate natural language instructions that demand long-context understanding, logical reasoning, and accurate mathematical computation. Two key attributes distinguish \textsc{RuleArena} from traditional rule-based reasoning benchmarks: (1) it extends beyond standard first-order logic representations, and (2) it is grounded in authentic, practical scenarios, providing insights into the suitability and reliability of LLMs for real-world applications. Our findings reveal several notable limitations in LLMs: (1) they struggle to identify and apply the appropriate rules, frequently becoming confused by similar but distinct regulations, (2) they cannot consistently perform accurate mathematical computations, even when they correctly identify the relevant rules, and (3) in general, they perform poorly in the benchmark. We also observe a significant performance boost when LLMs are provided with external tools for oracle math and logic operations. These results highlight significant challenges and promising research directions in advancing LLMs' rule-guided reasoning capabilities in real-life applications. Our codes and data are publicly available on \href{https://github.com/skyriver-2000/rulearena}{GitHub}.
\end{abstract}

\input{contents/1-intro}
\input{contents/2-related}
\input{contents/3-dataset}
\input{contents/4-exp}

\input{contents/5-conclusion}
\input{contents/6-limitation}

\bibliography{references}

\newpage
\onecolumn
\appendix
\input{contents/appendix}

\end{document}

%% file: math_commands.tex
\usepackage{amsmath,amsfonts,bm}

\def\eqref#1{equation~\ref{#1}}

\def\1{\bm{1}}

\DeclareMathAlphabet{\mathsfit}{\encodingdefault}{\sfdefault}{m}{sl}
\SetMathAlphabet{\mathsfit}{bold}{\encodingdefault}{\sfdefault}{bx}{n}



%% file: commands.tex
\newcommand{\se}[1]{Section~\ref{#1}}
\newcommand{\ap}[1]{Appendix~\ref{#1}}



%% file: contents/1-intro.tex
\section{Introduction}

\begin{figure*}[tbp]
    \centering
    \includegraphics[width=0.92\linewidth]{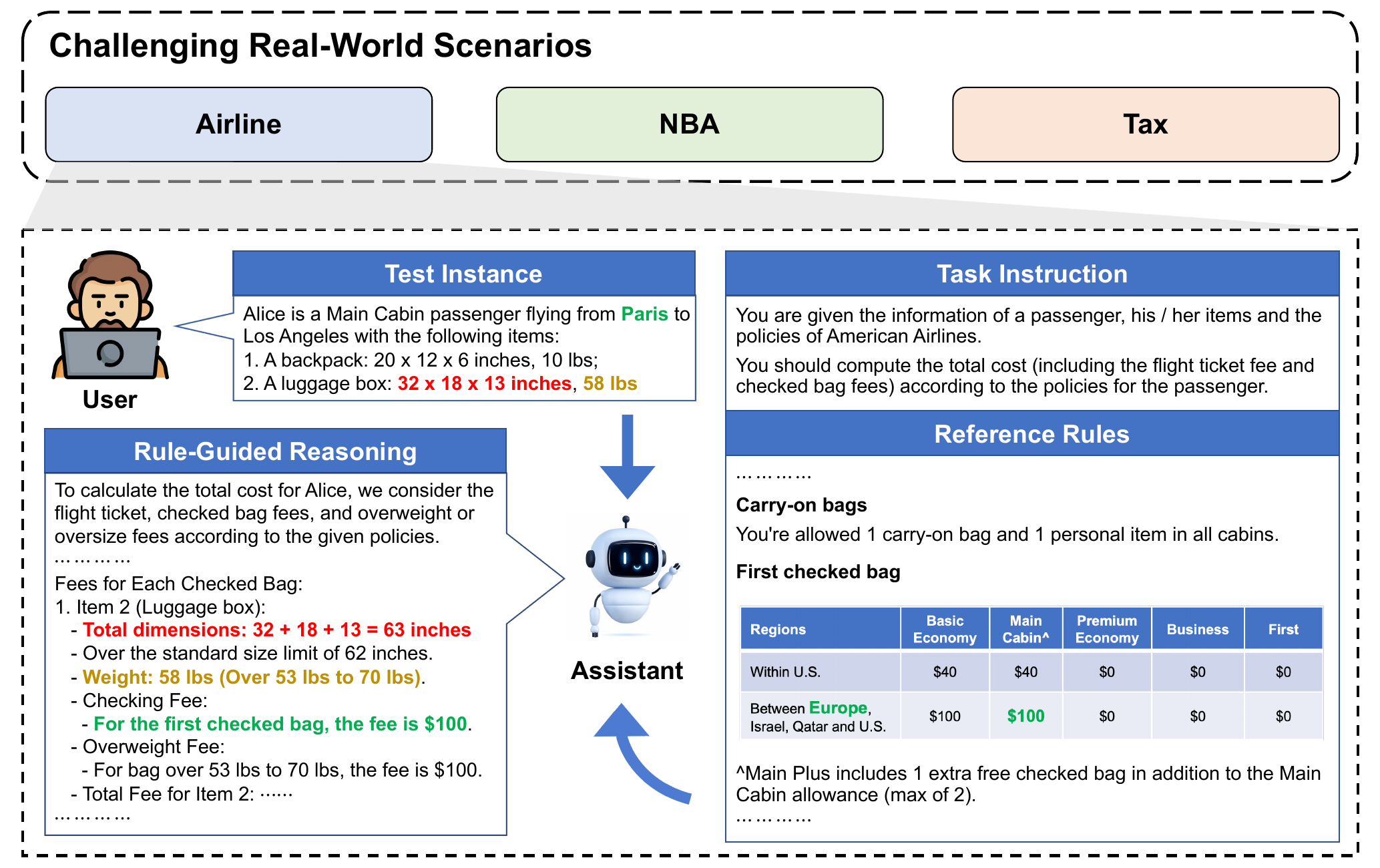}
    \vspace{-4pt}
    \caption{Overview of \textsc{RuleArena}. \textsc{RuleArena} contains 95 commonly used and moderately complex rules and 816 test problems from three representative real-world scenarios - airline luggage fees, NBA transactions, and taxation policies. LLMs are given a the task instruction, the reference rules in this scenario, and a user instance, and required to conduct reasoning and computation for the user input under the guidance of reference rules.}
    \label{fig:framework}
    \vspace{-8pt}
\end{figure*}

Recently, Large Language Models (LLMs) \citep{touvron2023llama,openai2023gpt4,google2023gemini,anthropic2024claude} have demonstrated remarkable capabilities across many real-world applications. However, their limited domain-specific knowledge often leads to unfaithful or misleading output, which can cause significant risks and financial liabilities. For example, Canadian airline was recently required to compensate a customer who received incorrect guidance from the airline’s chatbot\footnote{\url{https://www.theguardian.com/world/2024/feb/16/air-canada-chatbot-lawsuit}}. These challenges highlight the need for robust, real-world benchmarks that assess how faithfully and accurately LLMs can follow real-life instructions and adhere to relevant regulations, thereby ensuring reliable and safe outputs for deployment.

Although several studies have examined LLMs’ instruction-following abilities \citep{chen2024codieval,jiang2024followbench,wen2024complexbench}, they mainly focused on style constraints, such as the format \cite{zhou2023instruction}, length, or topic of responses. Yet, the significance of instruction-following extends well beyond style compliance. In many problem-solving scenarios, instructions function as \emph{rules}: they impose logical constraints on the reasoning process and specify how answers should be derived from given inputs. However, limited attention has been paid to LLMs' capacity to follow complex rules. 

Existing research \citep{mu2023rule,sun2024rulebench} largely addresses only single-step, first-order logic reasoning or artificially synthesized logical tasks. In contrast, real-world rules frequently appear in diverse and nuanced natural language forms. They may involve intricate logical structures, including the need for parallel reasoning across multiple rules or navigating interdependent rule sets. 
For instance, to calculate the fees for checked luggage when taking flights, one needs to consider the base price for checking each item, the overweight and oversize charges, and how these charges should be aggregated together. The extent to which LLMs can accurately follow these complex, real-world rules---an ability we term \emph{rule-guided reasoning}---remains unknown.
To better understand the complexity and practical implications of rule-guiding reasoning, we introduce a new evaluation benchmark, \textsc{RuleArena}, grounded in realistic scenarios.

As illustrated in Figure \ref{fig:framework}, \textsc{RuleArena} is developed from three representative, real-world scenarios: (1) airline luggage fee policies, (2) NBA transactions, and (3) taxation policies. From these domains, we collect authentic rules currently implemented by companies or government agencies. For each domain, we construct a set of challenging test problems, pairing each question with a ground-truth solution, and then evaluate a range of state-of-the-art LLMs on their ability to conform to the rules. LLMs are provided with the domain-specific task instructions and reference rules, and required to resolve each test problem through reasoning according to the question and reference rules.

Our contributions in this work can be summarized in three main points:

\begin{itemize}
    \item \vspace{-4pt}\textbf{A diverse collection of real-world rules:} We assemble a comprehensive set of 95 policies/rules drawn from these three real-world scenarios.
    \item \vspace{-4pt}\textbf{A challenging benchmark and novel evaluation metrics:} Using the collected rules, we introduce \textsc{RuleArena}, a new benchmark containing 816 datapoints designed to test LLMs' rule-guided reasoning ability. We further propose a suite of evaluation metrics for both rule selection and rule application, providing fine-grained insights into LLMs’ performance.
    \item \textbf{A comprehensive analysis of prevalent challenges:} By examining common failure cases, identifying difficult rule types, and conducting extensive controlled experiments, we uncover several systematic issues that limit current LLMs' rule-guided reasoning capabilities and promising directions to improve.
\end{itemize}

%% file: contents/2-related.tex
\section{Related Work}

\paragraph{Complex Instruction-following Benchmarks} A wide range of benchmarks has been designed to evaluate LLMs’ instruction-following abilities from various perspectives, including semantics \citep{zheng2023llmasajudge, li2023alpacaeval, liu2023alignbench,wu2024akew,wu2024antileak}, format \citep{xia2024fofo, tang2023struc}, and response length \citep{chen2024benchmarking, sun2023evaluating}. To further probe complexity, some works have introduced benchmarks that construct complex instructions through compositional methods. For example, \textbf{WizardLM} \citep{xu2023wizardlm} generates intricate tasks by combining simpler instructions, while \textbf{CELLO} \citep{he2024can} uses task descriptions and input texts to create complex prompts grounded in real-world scenarios. \textbf{ComplexBench} \citep{wen2024complexbench} adopts multiple compositional structures to integrate atomic requirements into more challenging instructions. In contrast, our work focuses on instructions derived directly from real-life scenarios, where naturally occurring complexities arise from multifaceted constraints incurred by inputs on the set of instructions.

\paragraph{Logical Reasoning Benchmarks}
Extensive research has explored benchmarks for mathematical \citep{kedziorski2016mawps,ling2017aqua,amini2019mathqa,cobbe2021gms8k,hendrycks2021math} and logical \citep{mao2019nscl,gupta2020nmn,tafjord2021proofwriter,zhong2022arlsat,han2022folio, zhang2024supervised} reasoning, evaluating LLMs' abilities to solve math problems of varying difficulty, tackle coding challenges, and engage in deductive logic. Although these benchmarks test models' reasoning skills, their logical constraints are often represented in simplified, formal systems, such as propositional \citep{hua2024disentangling} or first-order logic \citep{zhu2023dyval, mu2023rule,sun2024rulebench}. In contrast, our benchmark deals with rules that arise in natural language, capturing a richer, more realistic set of constraints. Such natural language rules extend beyond neatly formalized logical representations, often express higher-order logic and more intricate relationships than typical propositional or first-order logic formalizations.

%% file: contents/3-dataset.tex
\section{RuleArena}
In this section, we present the \textsc{RuleArena} benchmark and its construction process. We begin by describing the domains we have chosen and the corresponding regulations from which our rules are collected. We then describe how problems with varying difficulty levels are generated and how the ground-truth solutions are computed. Finally, we present the evaluation metrics we used to evaluate whether correct rules are correctly applied.

\subsection{Domains and Rule Collection}
We select three real-life domains that all both familiar in everyday life and demonstrate a high level of complexity:

\paragraph{Airline.} It requires LLM to calculate the total cost for one or more passengers, including their flight ticket and checked baggage fees. The regulations are extracted from policy of American Airlines\footnote{\url{https://www.aa.com/i18n/customer-service/support/optional-service-fees.jsp}}. The complexity stems from the fact that baggage costs vary according to factors such as cabin class, flight origin and destination, the number of checked bags, and the size of each bag. Consequently, LLMs must carefully identify the correct baggage-related rules and apply them accurately to determine the final cost.

\paragraph{NBA transaction.} It requires LLMs to determine whether one or more specified transactions are allowed. The regulations are extracted from the \emph{2023 NBA Collective Bargaining Agreements}\footnote{\url{https://ak-static.cms.nba.com/wp-content/uploads/sites/4/2023/06/2023-NBA-Collective-Bargaining-Agreement.pdf}} (CBA) and excerpt from the \emph{NBA Constitution and By-Laws}\footnote{\url{https://ak-static-int.nba.com/wp-content/uploads/sites/3/2015/12/NBA-Constitution-and-By-Laws.pdf}}. Complexity arises from the numerous factors influencing transaction eligibility, including the player’s contract value, salary-matching constraints, and the specific transaction date. LLMs must accurately identify and apply the relevant rules from the agreement to determine whether a given transaction can proceed.

\paragraph{Tax.} It requires LLMs to calculate the income tax for one person or family given their financial information. The regulations are collected from Internal Revenue Service\footnote{\url{https://www.irs.gov/forms-instructions}}. Although taxes are a common and universally encountered aspect of modern life, they are also known for their complexity. This complexity stems from a wide range of factors, including salary income, investment gains, gifts, home ownership and related expenses, as well as the jurisdiction in which income is earned. To arrive at the correct tax amount, LLMs must navigate and apply the appropriate rules drawn from these multifaceted conditions.

The statistics for the collected rules are summarized in Table~\ref{tab:rule-statistics}. Although the total number of rules is relatively small, each rule averages just under 400 tokens in length, tokenized by Llama-3.1 tokenizer. This presents a substantial challenge for both rule comprehension and the handling of long contexts. For more details on how we collect our rules, please refer to  \ap{ap:rules}.

\begin{table}[!t]
    \centering
    \resizebox{6cm}{!}{\begin{tabular}{cccc}
    \toprule
         & Airline & NBA& Tax \\
         \midrule
        \# Rules & 10 & 54 & 31 \\
        Average \# Tokens & 376 & 398 & 359 \\
        \bottomrule
    \end{tabular}}
    \vspace{-4pt}
    \caption{Statistics of rules in each domain.}
    \label{tab:rule-statistics}
    \vspace{-8pt}
\end{table}

\subsection{Problem Annotation}
After gathering the relevant rules for each domain, we construct challenging test problems designed to evaluate whether LLMs can produce correct outputs from the provided rules.

\paragraph{Airline.} The problems are generated by randomly selecting passenger information (e.g., cabin class, itinerary, ticket price) and their checked baggage details (e.g., dimensions, weight). We convert each regulation into a corresponding rule-based script, enabling the direct calculation of ground-truth answers by executing these scripts. LLM performance is then assessed by comparing the model’s computed solutions to the script-derived ground truths, step by step.

\paragraph{NBA Transaction.} The problems consist of proposed trades that may or may not comply with NBA  regulations. Because these problems require a wide variety of operations and rule sets, fully automated generation and evaluation are difficult. Therefore, we employ annotators familiar with NBA transaction rules to curate complex test cases and identify all the relevant rules needed to resolve each case (further details in Appendix \ref{ap:annotation}). For each problem, we ask the LLM whether the transaction is legit or not based on the regulations. If LLM thinks the transaction is legit, it should generate ``Yes''; otherwise, it needs to identify the specific team and transaction that violates the rules.

\paragraph{Tax.} The problems are randomly generated from hypothetical taxpayer profiles including information such as  income levels, filing status, \emph{etc}. IRS tax regulations are translated into rule-based scripts to compute ground-truth tax obligations. As with the airline scenario, we measure LLM accuracy by comparing the model’s step-by-step calculations with those derived directly from the scripts.

\subsection{Difficulty control}
To assess LLMs' capabilities under varying levels of complexity, we create problems with different degrees of difficulty. We define three levels of difficulties in each domain.

\paragraph{Airline.} The difficulty is controlled by adjusting the number of bags a passenger carries. 

\paragraph{NBA Transaction.} Complexity is determined by increasing the number of teams, players, and transactions involved in a scenario. 

\paragraph{Tax.} The level of difficulty is raised by progressively introducing additional tax forms and thus relevant regulations.

The statistics of problems at different difficulty levels in each domain are listed in Table~\ref{tab:task-statistics}.

\begin{table}[ht]
    \centering
    \resizebox{4.5cm}{!}{
    \begin{tabular}{cccc}
        \toprule
         & Airline & NBA & Tax \\
         \midrule
        Level 1 & 100 & 81 & 100 \\
        Level 2 & 100 & 89 & 100 \\
        Level 3 & 100 & 46 & 100 \\
        In Total & 300 & 216 & 300 \\
        \bottomrule
    \end{tabular}
    }
    \vspace{-4pt}
    \caption{Number of test problems at different difficulty levels in each domain.}
    \label{tab:task-statistics}
    \vspace{-8pt}
\end{table}

\subsection{Evaluation Metrics}\label{subsec:metric}

To achieve a comprehensive evaluation of the rule-following abilities of Large Language Models (LLMs), we introduce a set of evaluation metrics. Unlike existing benchmarks \citep{hua2024disentangling, fan2023nphardeval, zhu2023dyval}, which primarily rely on simple metrics such as answer accuracy or BLEU scores, our approach aims to conduct a more detailed analysis of the step-by-step rule-guided reasoning process. This analysis includes examining each rule application to determine whether the rule should be applied, whether any rules are missed, and whether the rule application computation process is accurate.

For each domain, assuming a set $\mathcal{T}$ of $N$ problems and a set $\mathcal{R}$ of $M$. For each problem $t_i=\left(q_i,a_i,R_i\right)$, we have a query $q_i$, an answer $a_i$, and a set of relevant rules $R_i$, together with a rule-usage matrix $U\in\mathbb{R}^{N\times M}$, where each item $U_{i,r}\in\{0,1\}$ indicates whether a rule $r$ is used by an LLM in problem $t_i$. Matrix $U$ can be approximately obtained by parsing LLMs' responses using an LLM, which we will introduce in \se{subsec:exp-setting}.

Now we introduce two groups of metrics:

\textbf{The first group focuses on problem-level evaluations}: for each problem, we examine whether all necessary rules were applied, whether any extraneous rules were applied, and whether the final answer aligns with the ground-truth solution:

\minisection{Problem-wise Recall}: denoted as $\rm R(t)$, measures whether LLMs apply all relevant rules for a problem $t$. For each problem $t_i$, $\rm P(t_i)$ is calculated as the proportion of relevant rules that are applied by LLMs:
\begin{equation}
    {\rm R}(t_i) = \frac{\sum_{r\in R_i}\mathbb{I}(U_{i,r}=1)}{\sum_r \mathbb{I}(r\in R_i)}
\end{equation}

\minisection{Problem-wise Rule Application Correctness}: denoted as $\rm AC(t)$, measures whether LLMs apply rules correctly for a problem $t$. For each problem $t_i$, $\rm AC$ is calculated as the proportion of correctly applied rules that are relevant:
\begin{equation}
    {\rm AC}(t_i) = \frac{\sum_{U_{i,r}=1}\mathbb{I}(r \text{ is correctly applied})}{\sum_r \mathbb{I}(U_{i,r}=1)}
\end{equation}

\minisection{Problem-wise Precision}: denoted as $\rm P(t)$, measures whether LLMs apply only relevant rules for a problem $t$. For each problem $t_i$, $\rm P(t_i)$ is calculated as the proportion of applied rules that are relevant:
\begin{equation}
    {\rm P}(t_i) = \frac{\sum_{U_{i,r}=1}\mathbb{I}(r\in R_i)}{\sum_r \mathbb{I}(U_{i,r}=1)}
\end{equation}

\minisection{Problem-wise Accuracy}: denoted as $\rm Acc(t)$, measures whether LLMs accurately answer the problem $t$ comparing with ground-truth result. Assume the LLM provides answer $\tilde{a}_i$ for a problem $t_i$, the accuracy should be calculated as:
\begin{equation}
    {\rm Acc}(t_i) = \mathbb{I}(\tilde{a}_i=a_i)
\end{equation}

\textbf{The second group of metrics focuses on rules rather than problems.} For each rule in the domain, we assess whether it is applied to all problems that require it, and whether the problems it is applied to are exactly those that necessitate it:

\minisection{Rule-wise Recall}, denoted as $R(r)$, measures whether LLMs decide to apply $r$ when $r$ is relevant to a problem:
\begin{equation}
    {\rm R}(r) = \frac{\sum_i\mathbb{I}(r\in R_i)\mathbb{I}(U_{i,r}=1)}{\sum_i\mathbb{I}(r\in R_i)}
\end{equation}

\minisection{Rule-wise Rule Application Correctness}, denoted as $AC(r)$, measures whether LLMs correctly apply $r$:
\begin{equation}
    {\rm AC}(r) = \frac{\sum_i\mathbb{I}(U_{i,r}=1)\mathbb{I}(r\text{ is correctly applied})}{\sum_i\mathbb{I}(U_{i,r}=1)}
\end{equation}

\minisection{Rule-wise Precision}, denoted as $P(r)$, measures whether $r$ is relevant to the problem when LLMs decide to apply $r$:
\begin{equation}
    {\rm P}(r) = \frac{\sum_i\mathbb{I}(U_{i,r}=1)\mathbb{I}(r\in R_i)}{\sum_i\mathbb{I}(U_{i,r}=1)}
\end{equation}

\emph{Problem Recall} and \emph{Problem Accuracy} are applied to all three domains to measure the ability of LLMs to match and aggregate rules and to comprehensively follow the rules. \emph{Problem Application Correctness} is used on Airline and Tax tasks to evaluate whether LLMs can operate correctly under the guidance of rules, as these two tasks have clear procedures without ambiguous rules, while \emph{Problem Precision} is used on NBA tasks to examine whether LLMs can differentiate similar rules applicable to different situations.

%% file: contents/4-exp.tex
\begin{table*}[t]
    \centering
    \resizebox{15cm}{!}{
    \begin{tabular}{l c cccccccccccc}
        \toprule
          \multirow{2}{*}{\bf Models}& \multirow{2}{*}{\bf Settings}& \multicolumn{4}{c}{\bf Level 1} & \multicolumn{4}{c}{\bf Level 2} & \multicolumn{4}{c}{\bf Level 3}\\
           \cmidrule(lr){3-6} \cmidrule(lr){7-10}  \cmidrule(lr){11-14}
          & & ${\rm P}(t)$  & ${\rm AC}(t)$ & ${\rm R}(t)$ & ${\rm Acc}(t)$ & ${\rm P}(t)$  & ${\rm AC}(t)$ & ${\rm R}(t)$ & ${\rm Acc}(t)$ & ${\rm P}(t)$  & ${\rm AC}(t)$ & ${\rm R}(t)$ & ${\rm Acc}(t)$ \\
         \midrule \multicolumn{14}{c}{\textbf{Airline}} \\
         \midrule
           \multirow{2}{*}{Llama-3.1 70B} & 0-shot & 1.000 & 0.764 & 0.558 & 0.01 & 1.000 & 0.732 & 0.535 & 0.01 & 1.000 & 0.752 & 0.578 & 0.00 \\
           & 1-shot & 1.000 & 0.809 & 0.787 & 0.17 & 1.000 & 0.827 & 0.801 & 0.07 & 1.000 & 0.769 & 0.815 & 0.01 \\
           \hdashline
           \multirow{2}{*}{Qwen-2.5 72B} & 0-shot & 1.000 & 0.636 & 0.586 & 0.01 & 1.000 & 0.627 & 0.554 & 0.01 & 1.000 & 0.588 & 0.544 & 0.00 \\
           & 1-shot & 1.000 & 0.836 & 0.908 & 0.19 & 1.000 & 0.818 & \colorbox{mylightblue}{0.901} & 0.10 & 1.000 & 0.801 & 0.904 & 0.01 \\
           \hdashline
           \multirow{2}{*}{Llama-3.1 405B} & 0-shot & 1.000 & 0.854 & 0.604 & 0.03 & 1.000 & 0.844 & 0.587 & 0.06 & 1.000 & 0.845 & 0.570 & 0.01 \\
           & 1-shot & 1.000 & 0.919 & \colorbox{mywheat}{\textbf{0.921}} & 0.32 & 1.000 & 0.897 & \colorbox{mywheat}{\textbf{0.905}} & 0.16 & 1.000 & 0.870 & \colorbox{mywheat}{\textbf{0.946}} & 0.04 \\
           \hdashline
           \multirow{2}{*}{Claude-3.5 Sonnet} & 0-shot & 1.000 & 0.930 & 0.702 & 0.04 & 1.000 & 0.876 & 0.669 & 0.00 & 1.000 & 0.888 & 0.646 & 0.01 \\
           & 1-shot & 1.000 & 0.960 & 0.871 & 0.29 & 1.000 & \colorbox{mywheat}{\textbf{0.966}} & 0.822 & 0.30 & 1.000 & \colorbox{mywheat}{\textbf{0.972}} & 0.718 & 0.11 \\
           \hdashline
           \multirow{2}{*}{GPT-4o} & 0-shot & 1.000 & 0.862 & 0.616 & 0.02 & 1.000 & 0.868 & 0.578 & 0.00 & 1.000 & 0.813 & 0.548 & 0.00 \\
           & 1-shot & 1.000 & 0.922 & 0.885 & 0.32 & 1.000 & 0.875 & 0.853 & 0.16 & 1.000 & 0.835 & 0.798 & 0.05 \\
           \hdashline
           \multirow{2}{*}{o1-preview} & 0-shot & 1.000 & \colorbox{mylightblue}{0.968} & 0.888 & \colorbox{mylightblue}{0.54} & 1.000 & 0.950 & 0.881 & \colorbox{mylightblue}{0.37} & 1.000 & 0.958 & 0.855 & \colorbox{mylightblue}{0.21} \\
           & 1-shot & 1.000 & \colorbox{mywheat}{\textbf{0.971}} & \colorbox{mylightblue}{0.911} & \colorbox{mywheat}{\textbf{0.63}} & 1.000 & \colorbox{mylightblue}{0.963} & \colorbox{mylightblue}{0.901} & \colorbox{mywheat}{\textbf{0.55}} & 1.000 & \colorbox{mylightblue}{0.961} & \colorbox{mylightblue}{0.929} & \colorbox{mywheat}{\textbf{0.46}} \\
           \midrule \multicolumn{14}{c}{\textbf{NBA Transaction}} \\
         \midrule
           \multirow{2}{*}{Llama-3.1 70B} & 0-shot & 0.579 & -- & 0.428 & 0.40 & 0.498 & -- & 0.246 & 0.36 & 0.540 & -- & 0.250 & 0.22 \\
           & 1-shot & 0.560 & -- & \colorbox{mywheat}{\textbf{0.565}} & 0.49 & 0.466 & -- & 0.386 & 0.25 & 0.578 & -- & \colorbox{mylightblue}{0.438} & 0.26 \\
           \hdashline
           \multirow{2}{*}{Qwen-2.5 72B} & 0-shot & 0.556 & -- & 0.409 & 0.44 & 0.537 & -- & 0.339 & \colorbox{mylightblue}{0.43} & 0.592 & -- & 0.305 & \colorbox{mywheat}{\textbf{0.30}} \\
           & 1-shot & 0.595 & -- & 0.526 & 0.53 & 0.495 & -- & 0.378 & 0.35 & 0.574 & -- & 0.327 & 0.17 \\
           \hdashline
           \multirow{2}{*}{Llama-3.1 405B} & 0-shot & 0.581 & -- & 0.419 & 0.49 & 0.577 & -- & 0.323 & 0.30 & 0.561 & -- & 0.297 & \colorbox{mylightblue}{0.28} \\
           & 1-shot & 0.608 & -- & \colorbox{mylightblue}{0.550} & \colorbox{mylightblue}{0.56} & 0.559 & -- & \colorbox{mylightblue}{0.439} & 0.29 & 0.575 & -- & \colorbox{mywheat}{\textbf{0.461}} & 0.10 \\
           \hdashline
           \multirow{2}{*}{Claude-3.5 Sonnet} & 0-shot & 0.660 & -- & 0.457 & 0.38 & 0.630 & -- & 0.373 & 0.40 & 0.588 & -- & 0.292 & \colorbox{mylightblue}{0.28} \\
           & 1-shot & 0.676 & -- & 0.528 & \colorbox{mywheat}{\textbf{0.58}} & 0.676 & -- & 0.410 & \colorbox{mywheat}{\textbf{0.47}} & 0.650 & -- & 0.371 & 0.26 \\
           \hdashline
           \multirow{2}{*}{GPT-4o} & 0-shot & 0.650 & -- & 0.446 & 0.40 & 0.570 & -- & 0.327 & 0.26 & 0.603 & -- & 0.291 & 0.24 \\
           & 1-shot & 0.616 & -- & 0.506 & 0.40 & 0.597 & -- & 0.392 & 0.28 & 0.569 & -- & 0.318 & 0.20 \\
           \hdashline
           \multirow{2}{*}{o1-preview} & 0-shot & \colorbox{mywheat}{\textbf{0.742}} & -- & 0.502 & 0.44 & \colorbox{mylightblue}{0.707} & -- & 0.430 & \colorbox{mywheat}{\textbf{0.47}} & \colorbox{mywheat}{\textbf{0.747}} & -- & 0.415 & 0.24 \\
           & 1-shot & \colorbox{mylightblue}{0.731} & -- & \colorbox{mywheat}{\textbf{0.565}} & \colorbox{mywheat}{\textbf{0.58}} & \colorbox{mywheat}{\textbf{0.715}} & -- & \colorbox{mywheat}{\textbf{0.512}} & 0.40 & \colorbox{mylightblue}{0.724} & -- & 0.413 & 0.20 \\
           \midrule \multicolumn{14}{c}{\textbf{Tax}} \\
         \midrule
           \multirow{2}{*}{Llama-3.1 70B} & 0-shot & 1.000 & 0.834 & 0.989 & 0.01 & 1.000 & 0.767 & 0.918 & 0.00 & 1.000 & 0.745 & 0.852 & 0.00 \\
           & 1-shot & 1.000 & 0.923 & 0.998 & 0.11 & 1.000 & 0.895 & 0.941 & 0.00 & 1.000 & 0.873 & 0.910 & 0.00 \\
           \hdashline
           \multirow{2}{*}{Qwen-2.5 72B} & 0-shot & 1.000 & 0.888 & 0.998 & 0.10 & 1.000 & 0.835 & 0.944 & 0.01 & 1.000 & 0.785 & 0.903 & 0.00 \\
           & 1-shot & 1.000 & 0.931 & \colorbox{mywheat}{\textbf{1.000}} & 0.17 & 1.000 & 0.919 & 0.934 & 0.00 & 1.000 & 0.921 & 0.868 & 0.00 \\
           \hdashline
           \multirow{2}{*}{Llama-3.1 405B} & 0-shot & 1.000 & 0.923 & \colorbox{mylightblue}{0.999} & 0.16 & 1.000 & 0.876 & 0.964 & 0.02 & 1.000 & 0.797 & 0.926 & 0.00 \\
           & 1-shot & 1.000 & 0.941 & \colorbox{mywheat}{\textbf{1.000}} & 0.24 & 1.000 & 0.914 & 0.958 & 0.03 & 1.000 & 0.873 & 0.880 & 0.00 \\
           \hdashline
           \multirow{2}{*}{Claude-3.5 Sonnet} & 0-shot & 1.000 & 0.964 & \colorbox{mywheat}{\textbf{1.000}} & 0.32 & 1.000 & 0.934 & 0.940 & 0.02 & 1.000 & 0.887 & 0.866 & 0.00 \\
           & 1-shot & 1.000 & 0.979 & \colorbox{mywheat}{\textbf{1.000}} & 0.64 & 1.000 & 0.954 & 0.969 & 0.16 & 1.000 & 0.895 & 0.888 & 0.00 \\
           \hdashline
           \multirow{2}{*}{GPT-4o} & 0-shot & 1.000 & 0.965 & \colorbox{mywheat}{\textbf{1.000}} & 0.42 & 1.000 & 0.951 & 0.957 & 0.07 & 1.000 & \colorbox{mylightblue}{0.945} & 0.908 & 0.00 \\
           & 1-shot & 1.000 & 0.975 & \colorbox{mywheat}{\textbf{1.000}} & 0.57 & 1.000 & \colorbox{mywheat}{\textbf{0.975}} & 0.944 & 0.07 & 1.000 & \colorbox{mywheat}{\textbf{0.982}} & 0.893 & 0.00 \\
           \hdashline
           \multirow{2}{*}{o1-preview} & 0-shot & 1.000 & \colorbox{mylightblue}{0.992} & \colorbox{mywheat}{\textbf{1.000}} & \colorbox{mywheat}{\textbf{0.72}} & 1.000 & 0.945 & \colorbox{mylightblue}{0.981} & \colorbox{mylightblue}{0.28} & 1.000 & 0.914 & \colorbox{mywheat}{\textbf{0.976}} & \colorbox{mylightblue}{0.19} \\
           & 1-shot & 1.000 & \colorbox{mywheat}{\textbf{0.994}} & \colorbox{mywheat}{\textbf{1.000}} & \colorbox{mylightblue}{0.68} & 1.000 & \colorbox{mylightblue}{0.960} & \colorbox{mywheat}{\textbf{0.994}} & \colorbox{mywheat}{\textbf{0.33}} & 1.000 & 0.894 & \colorbox{mylightblue}{0.939} & \colorbox{mywheat}{\textbf{0.24}} \\
           \bottomrule
    \end{tabular}
    }
    \vspace{-4pt}
    \caption{Main problem-wise evaluation results on airline, NBA, and tax domains. ${\rm P}(t)$ denotes problem-wise precision, ${\rm AC}(t)$ denotes problem-wise rule application correctness, and ${\rm R}(t)$ denotes problem-wise recall. The best and second best results in each column are rendered with orange and blue backgrounds respectively.}
    \label{tab:main_table}
    \vspace{-10pt}
\end{table*}

\section{Experiments}\label{sec:exp}
This section presents the experiments on benchmark. We first introduce the LLMs and prompting strategies we use to evaluate, and then present the evaluation result.

\subsection{Experiment Settings}\label{subsec:exp-setting}

\minisection{LLMs.} Our \emph{rules}, which are prompted directly into LLMs, can be of a length up to 20,000 tokens. Therefore, we only consider LLMs that can handle such long contexts, including Llama-3.1 70B, Llama-3.1 405B \citep{dubey2024llama3}, Qwen-2.5 72B \citep{qwen2.5}, Claude-3.5 Sonnet \citep{anthropic2024claude}, GPT-4o \citep{openai2024gpt4o}, and o1-preview \citep{openai2024o1}.

\minisection{Prompting Strategies.} Since rule-guided reasoning can be an intricate multi-step reasoning process in our three real-world scenarios, we use Chain-of-Thought (CoT) \citep{wei2022cot,kojima2022zeroshotcot} reasoning by default. To further study if LLMs can learn to follow hard rules through in-context examples, we also compare 0-shot with 1-shot CoT given an example including a task of the lowest difficulty and its solution. Due to context limit, we do not further increase the number of in-context examples.

\minisection{Output Parsing.} To obtain the rule-usage matrix $U$ we mentioned in \se{subsec:metric}, we utilize the structured output mode of GPT-4o \citep{openai2024gpt4o} to parse the raw textual responses from LLMs. Specifically, for airline and tax problems we structuralize the ground-truth calculation process and ask GPT-4o to fill in problem-specific information according to an LLM's response, while for NBA problems we enumerate a list of all rules and ask GPT-4o to directly judge whether a specific rule is applied in an LLM's response. For details we refer our readers to \ap{ap:parse}.

\subsection{Main Results}\label{subsec:exp-main}
This section provides a comprehensive analysis of benchmark results. The analysis is divided into two parts: problem-wise analysis and rule-wise analysis. The problem-wise analysis evaluates the performance of LLMs across different problems and difficulty levels; the rule-wise analysis delves into how effectively LLMs identify and apply specific rules, highlighting common failure modes and the impact of rule complexity and similarity. 

\subsubsection{Problem-wise Analysis} Table~\ref{tab:main_table} presents the evaluation results\footnote{In NBA domain, the problem-wise correctness (${\rm AC}(t)$) of rule application could not be computed due to the absence of step-by-step computation annotations. Generating such detailed annotations would require extensive human effort.}. Notice that the values of precision (${\rm P})(t)$, rule application (${\rm AC}(t)$), and recall (${\rm R}(t)$) are much higher than accuracy (${\rm Acc}(t)$). This is because solving a problem requires using multiple rules, hence one correct rule recall or application is insufficient for a correct answer. For example, if a problem requires 10 rules and only one rule is missed, ${\rm R}(t)$ is high as 0.9 while very probably leading to mistaken final answer (${\rm Acc}(t)=0$).

\minisection{Low Accuracy.} Overall performance in problem result accuracy (${\rm Acc}(t)$), as summarized in Table~\ref{tab:main_table}, remains unsatisfactory across all three scenarios. Under the 0-shot setting, non-reasoning LLMs such as Llama 405B, Claude-3.5, and GPT-4o fail to produce correct answers for the simplest test problems, and even advanced reasoning model like o1-preview can solve only about 50\%$\sim$60\% of Level 1 problems. For more challenging problems, particularly in the airline and tax domains, ${\rm Acc}(t)$ of non-reasoning models rarely exceeds 10\% and o1-preview fails most of them as well. In 1-shot setting, we notice marked improvements on the easiest problems, yet the gains diminish as problem difficulty increases. These persistently low ${\rm Acc}(t)$ highlight the inherent complexity of the \textsc{RuleArena} benchmark and emphasize the need for more robust LLM reasoning and rule-following capabilities.

\minisection{High Precision.} In both the airline and tax scenarios, LLMs achieve 100\% precision (${\rm P}(t)$) in rule selection precision, consistently applying only those rules that are required. We notice that high ${\rm P}(t)$ stems from the relative clarity of the rules in these domains; the rules are neither highly similar nor ambiguous, making it straightforward to determine which ones apply. In contrast, the NBA scenario presents a more challenging environment, leading to noticeably lower ${\rm P}(t)$ in rule selection.

\minisection{Low Recall.} Despite exhibiting high ${\rm P}(t)$ in certain domains, LLMs often struggle with rule \textit{recall} (${\rm R}(t)$). Low ${\rm R}(t)$ in the airline, NBA, and more complex tax problems indicate that models do not fully grasp the reasoning workflows required. Consequently, they frequently fail to recall all necessary rules, reflecting an incomplete or superficial understanding of the underlying logic.

\minisection{High Rule Application Correctness.} While LLMs demonstrate relatively high application correctness (${\rm AC}(t)$) on rule application computation on average, ${\rm AC}(t)$ never reaches a perfect 100\%. Occasional errors in mathematical calculations or logical operations emerge even under explicit rule guidance. Although these mistakes are not pervasive, a single computational error can significantly compromise the final output’s accuracy (${\rm Acc}(t)$) in many cases. This observation underscores the importance of improving the reliability of math computation abilities in LLMs.

\begin{figure}[!ht]
    \centering
    \begin{subfigure}[b]{0.44\textwidth}
        \includegraphics[width=\columnwidth]{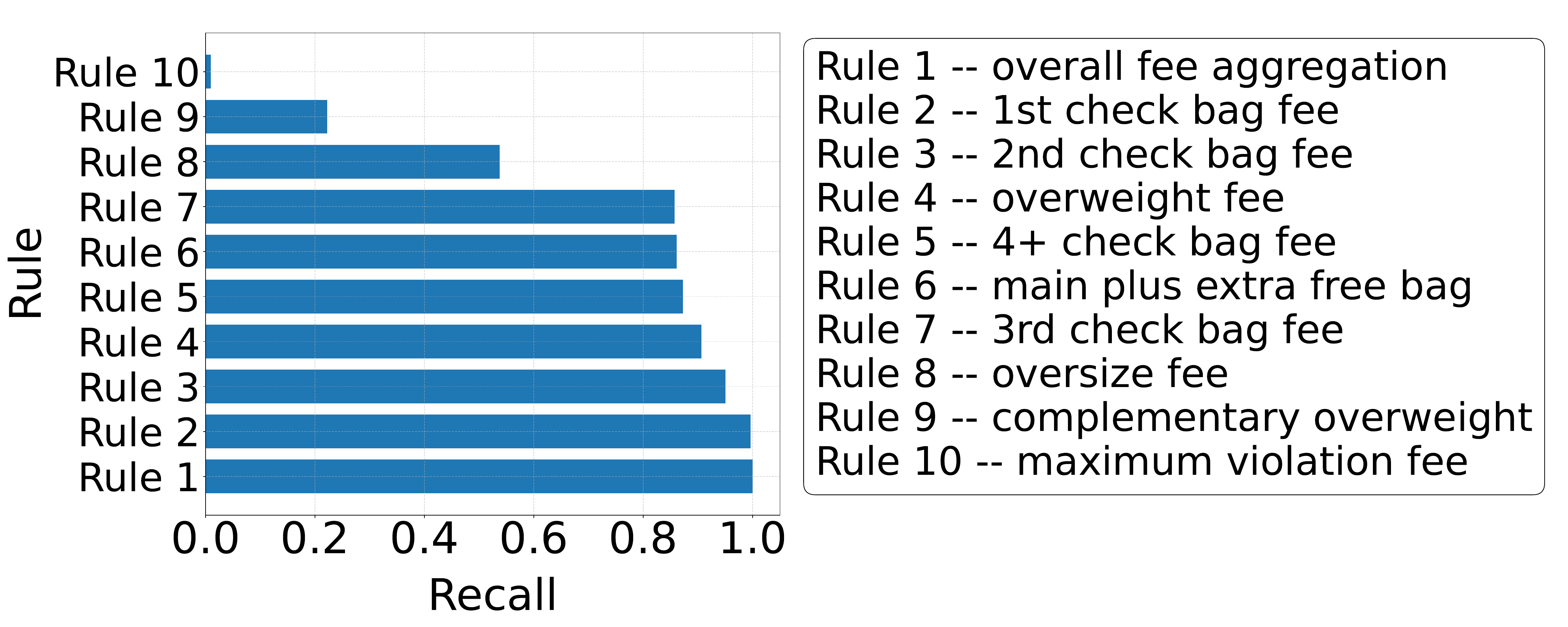}
        \caption{Recall}
        \label{fig:rule_wise_recall_airline}
    \end{subfigure}\\
    \begin{subfigure}[b]{0.44\textwidth}
        \includegraphics[width=\columnwidth]{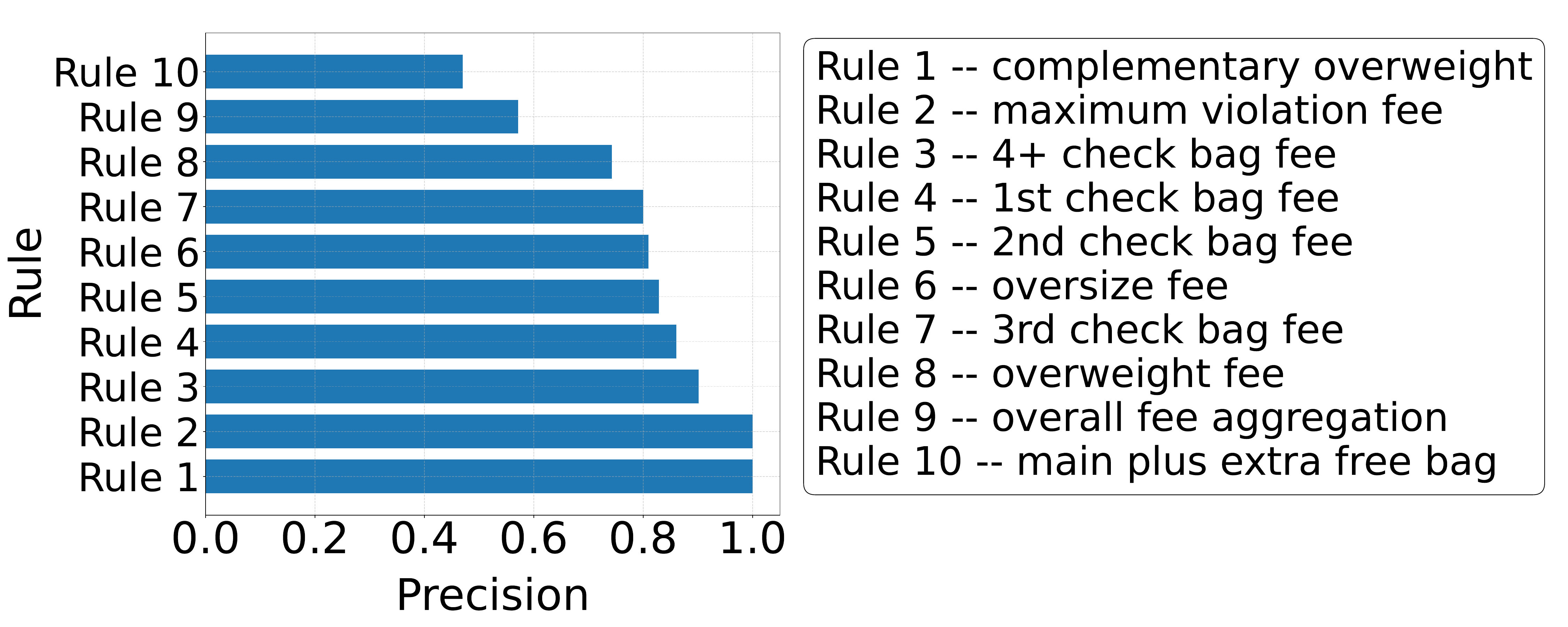}
        \caption{Correctness}
        \label{fig:rule_wise_correctness_airline}
    \end{subfigure}
    \caption{Rule-wise metrics of rules in airline domain.}
    \label{fig:rule_wise_airline}
\end{figure}

\subsubsection{Rule-wise Analysis}
Here, we provide a detailed examination of the rule-level evaluation. 
Figure \ref{fig:rule_wise_airline} presents recall (${\rm R}(r)$) and application correctness (${\rm AC}(r)$) in airline domain and metrics for NBA transaction and tax domains are presented in Figure \ref{fig:rule_wise_nba} and Figure \ref{fig:rule_wise_tax} in Appendix. Table~\ref{tab:stat_rule_wise} summarizes the metric results by reporting the mean and variance of three key metrics across all rules: recall (${\rm R}(r)$), application correctness (${\rm AC}(r)$), and precision (${\rm P}(r)$). The low variance observed in metrics such as ${\rm P}(r)$ within the airline and tax domains suggests that certain performance aspects are largely independent of the specific rules being applied. In contrast, the high variance seen in metrics like ${\rm R}(r)$ implies that recall performance is significantly influenced by the particular rules in question. Detailed analysis is presented below in Table~\ref{tab:stat_rule_wise}.

\begin{table}[htbp]
    \centering
    \resizebox{0.8\columnwidth}{!}{
    \begin{tabular}{lccc}
    \toprule
         & Airline & NBA & Tax \\
    \midrule
        ${\rm Mean}({\rm P}(r))$ & 1.000 & 0.504 & 1.000 \\
        ${\rm Var}({\rm P}(r))$ & 0.000 & 0.110 & 0.000 \\
    \hdashline
        ${\rm Mean}({\rm Ac}(r))$ & 0.798 & -- & 0.828 \\
        ${\rm Var}({\rm Ac}(r))$ & 0.026 & -- & 0.047 \\
    \hdashline
        ${\rm Mean}({\rm R}(r))$ & 0.721 & 0.308 & 0.900 \\
        ${\rm Var}({\rm R}(r))$ & 0.109 & 0.082 & 0.050 \\
    \bottomrule
    \end{tabular}
    }
    \caption{Statistics of our three rule-wise metrics.}
    \label{tab:stat_rule_wise}
\end{table}

\minisection{Rules with Low Recall.} Certain rules are systematically overlooked across multiple data points, indicating that their neglect is not random but concentrated on specific rules. To understand which rules are most frequently disregarded, we identify the top-5 rules with the lowest \emph{recall} (${\rm R}(r)$), as presented in Table~\ref{tab:rule_low_recall}. We find that most of these rules are ``non-essential,'' meaning they apply only under specific conditions. In contrast, ``essential'' rules must be applied in every scenario. For example, in the airline domain, essential rules define the baseline costs for each piece of luggage and the flight itself, making them relevant to all situations. Conversely, rules pertaining to overweight or oversized baggage only apply when such conditions arise, rendering them non-essential. Our observations indicate that these scenario-dependent, non-essential rules are more frequently neglected.

\begin{table*}[!ht]
    \centering
    \resizebox{\linewidth}{!}{
    \begin{tabular}{lclclc}
        \toprule
        \multicolumn{2}{c}{Airline} & \multicolumn{2}{c}{NBA} & \multicolumn{2}{c}{Tax} \\
        \cmidrule(lr){1-2} \cmidrule(lr){3-4} \cmidrule(lr){5-6}
        \multicolumn{1}{c}{Rule} & \multicolumn{1}{c}{essential} & \multicolumn{1}{c}{Rule} & \multicolumn{1}{c}{essential} & \multicolumn{1}{c}{Rule} & \multicolumn{1}{c}{essential} \\
        \midrule
        maximum violation fee & & salary space consumption of bird right & & education credits & \\
        complementary overweight & & salary space consumption of early bird right & & american opportunity credit & \\
        oversize fee & & sign and trade maximum salary & $\boldsymbol{\checkmark}$ & net profit & \\
        3rd base check fee & $\boldsymbol{\checkmark}$ & Arenas provision & & ctc or other dependent credit & \\
        main plus extra free bag & & over 38 rule & & taxes with qualified dividends & $\boldsymbol{\checkmark}$ \\
        \bottomrule
    \end{tabular}
    }
    \caption{Top-$5$ rules of the lowest \emph{recall} in ascent order of \emph{recall}.}
    \label{tab:rule_low_recall}
\end{table*}

\begin{table*}[t]
    \centering
    \resizebox{12cm}{!}{
    \begin{tabular}{lclc}
        \toprule
        \multicolumn{2}{c}{Airline} & \multicolumn{2}{c}{Tax} \\
        \cmidrule(lr){1-2} \cmidrule(lr){3-4}
        \multicolumn{1}{c}{Rule} & \multicolumn{1}{c}{Composition} & \multicolumn{1}{c}{Rule} & \multicolumn{1}{c}{Composition} \\
        \midrule
        main plus extra free bag & $\boldsymbol{\checkmark}$ & taxes with qualified dividends & $\boldsymbol{\checkmark}$ \\
        overall fee aggregation & $\boldsymbol{\checkmark}$ & standard taxes & \\
        overweight fee matching & & itemized deductions & $\boldsymbol{\checkmark}$ \\
        3rd base check fee & & standard deductions & $\boldsymbol{\checkmark}$ \\
        oversize fee matching & $\boldsymbol{\checkmark}$ & total income & $\boldsymbol{\checkmark}$ \\
        \bottomrule
    \end{tabular}
    }
    \caption{Top-$5$ rules of the lowest \emph{correctness} in ascent order of \emph{correctness}.}
    \label{tab:rule_low_correctness}
\end{table*}

\minisection{Rule with Low Application Correctness.} We also identified the top-5 rules with the lowest \emph{correctness} (${\rm AC}(r)$), listed in Table~\ref{tab:rule_low_correctness}. The majority of these rules are ``compositional'' in nature, requiring the aggregation of at least two previously computed intermediate results. By contrast, ``non-compositional'' rules demand at most a single mathematical operation involving a single intermediate result. Our analysis shows that compositional rules yield significantly lower ${\rm AC}(r)$ scores, indicating that LLMs struggle more with problems involving multiple reasoning steps than with straightforward, one-step computations.

\minisection{Rules with Low Precision.} In both the airline and tax scenarios, all rules exhibit high \textit{precision} (${\rm P}(r)$), indicating that LLMs rarely apply irrelevant rules during the reasoning process. However, the NBA domain presents a different challenge, where multiple rules appear similar. As shown in Table~\ref{tab:rule_low_precision}, rules with low precision in the NBA domain usually have alternatives applicable under different conditions in the same situation. This pattern suggests that when rules are easily confused with one another, LLMs struggle to consistently identify and apply the correct one.

\begin{table}[htbp]
    \centering
    \resizebox{\columnwidth}{!}{
    \begin{tabular}{lc}
    \toprule
        \multicolumn{1}{c}{\textbf{Rule}} & \textbf{Substitutable} \\
    \midrule
        higher max criterion & \\
        non bird right & $\boldsymbol{\checkmark}$ \\
        taxpayer mid level exception hard cap & $\boldsymbol{\checkmark}$ \\
        standard traded player exception & $\boldsymbol{\checkmark}$ \\
        salary increase ratio except bird right & $\boldsymbol{\checkmark}$ \\
    \bottomrule
    \end{tabular}
    }
    \caption{Top-$5$ rules of the lowest \emph{precision} in ascent order of \emph{precision}.}
    \label{tab:rule_low_precision}
\end{table}

\subsection{In-Depth Analyses}

\begin{figure*}[tbp]
    \centering
    \includegraphics[width=\linewidth]{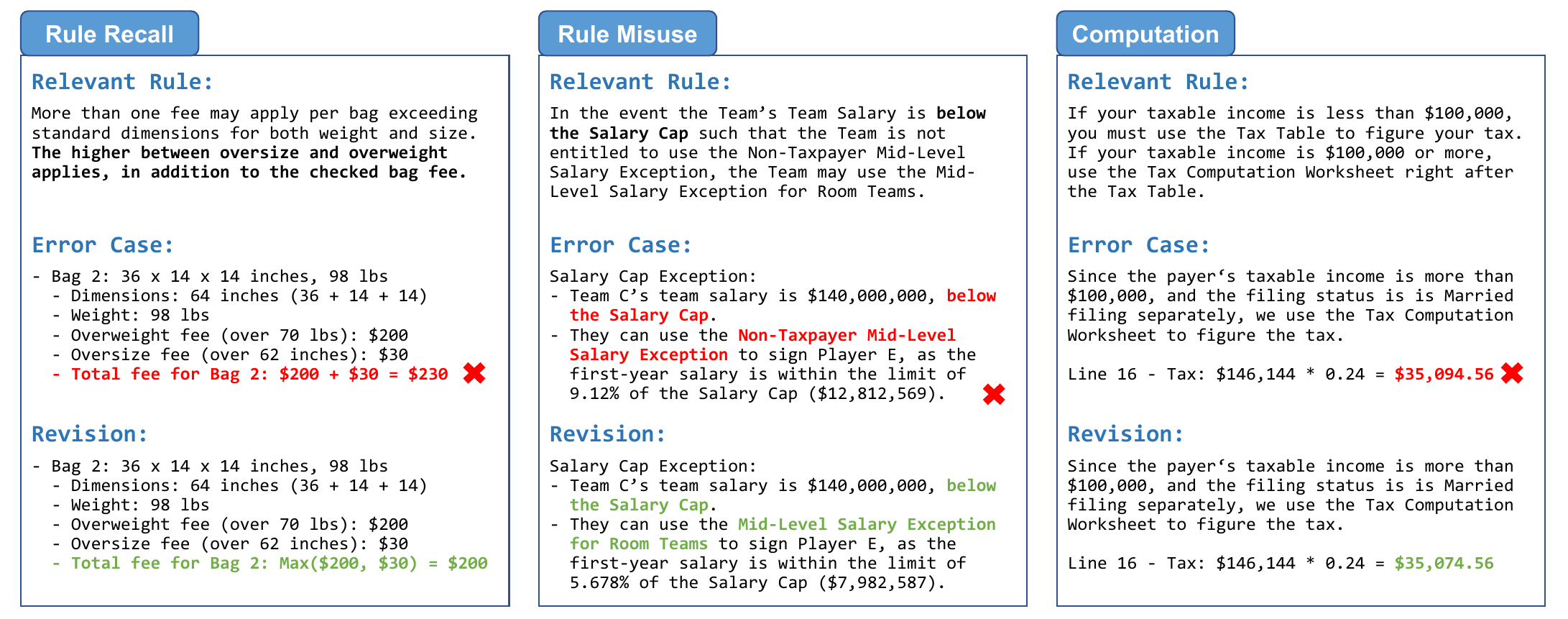}
    \caption{Failure Case Studies. Existing LLMs commonly fail due to inadequate rule recall, inappropriate usage of similar rules, and computation errors.}
    \vspace{-4pt}
    \label{fig:case_study}
    \vspace{-8pt}
\end{figure*}

\subsubsection{What Impacts Rule Following?}
We study the factors influencing LLM performance, as measured by ${\rm Acc}(t)$, and provide the complete experiment results in \ap{ap:rule_following_factor}. The main findings are:

\minisection{\hyperref[ap:corr-between-metrics]{Correlation between ${\rm Acc}(t)$ and other metrics.}} We compare the correlation between problem-wise metrics (i.e., ${\rm P}(t)$, ${\rm AC}(t)$, ${\rm R}(t)$) and accuracy ${\rm Acc}(t)$. The correlation is the most obvious and almost linear between ${\rm R}(t)$ and ${\rm Acc}(t)$, while highly non-linear or unclear between other two metrics and ${\rm Acc}(t)$.

\minisection{\hyperref[ap:in-context-example]{The effect of in-context examples.}} We observe that LLMs generally provide better performances given 1-shot example on airline, tax, and (easy) NBA problems. However, when tackling more challenging NBA problems (Levels 2 and 3), providing an example increases ${\rm P}(t)$ and ${\rm R}(t)$ but leads to a counterintuitive decrease in overall ${\rm Acc}(t)$.

\minisection{\hyperref[ap:rule-representation]{Rule representation has a mild effect.}} In the airline and tax domains, some rules are represented as Markdown tables. To test whether representation format affects performance, we convert these tabular rules into textual ``if-then'' statements and compare with original results. The comparison shows that converting tabular rules into text improves ${\rm R}(r)$, but has little impact on other metrics, including ${\rm Acc}(t)$.

\minisection{\hyperref[ap:distractive-rules]{Distractive rules degrades LLM performance.}} An essential aspect of rule-following involves identifying which rules are relevant to the current problem. We assess the extent to which irrelevant rules detrimentally affect performance, and notice that the presence of distractive (irrelevant) rules significantly degrades LLM performance.

\minisection{\hyperref[ap:tool-augmentation]{Tool augmentation boosts overall performance.}} Using external tools is a simple way to reduce math and logic errors from LLMs. To study to what extent external tools can help LLM rule-guided reasoning, we ask our LLMs to write Python code and use the execution result as solution, where Python interpreter serves as an oracle math and logic calculator. We observe that LLMs can achieve a significant performance boost with tool augmentation but are still far from perfect.

\subsubsection{Case Studies}

To gain an intuitive understanding of how and why LLMs fail in complex rule-following problems, we present representative failure cases in Figure \ref{fig:case_study}. These examples highlight three frequently observed failure modes:

\minisection{LLMs fail to recall certain rules.} As discussed in \se{subsec:exp-main}, LLMs often neglect non-essential rules. In airline problems, for instance, a crucial requirement is to apply either the oversize fee or the overweight fee (whichever is higher) and not to sum them. However, LLMs frequently overlook this instruction and incorrectly combine both fees, resulting in an inflated, incorrect total cost.

\minisection{LLMs get confused by similar rules.} When multiple rules appear similar but are applicable under different conditions, LLMs can misapply them. For example, in the NBA domain, teams under the Salary Cap should use the Mid-Level Exception for Room Teams, whereas teams above the Salary Cap should apply the Non-Taxpayer Mid-Level Exception. As illustrated in the second failure case of Figure \ref{fig:case_study}, LLMs sometimes conflate these exceptions. Similar confusion also arises with various Traded Player Exceptions and differing types of Bird Rights.\footnote{Explanations of these specific NBA terms can be found in the \ap{ap:NBA_terms}.}

\minisection{LLMs compute incorrect results.} Mathematical and logical operations present ongoing challenges. For example, in the tax scenario, LLMs must accurately compute a series of values related to income, tax brackets, and credits. Even a minor arithmetic mistake compromises the final result, as shown in the third failure case. Such computational errors underscore the need for more precise and reliable reasoning capabilities in LLMs.

%% file: contents/5-conclusion.tex
\section{Conclusions}

In this paper, we introduce \textsc{RuleArena}, a real-world benchmark designed to evaluate the abilities of LLMs on various rule-guided reasoning tasks. We observe that existing LLMs face significant challenges when they try to tackle problems on \textsc{RuleArena} - even the strongest Claude-3.5 and GPT-4o models can hardly succeed on our hardest tasks. Our further analysis indicates that LLMs struggle to integrate multiple rules or facts cohesively and are prone to irrelevant distractions.
\textsc{RuleArena} poses fundamental challenges in complex rule following, and provides a valuable tool for understanding and enhancing the reasoning abilities of LLMs. We envision \textsc{RuleArena} as a foundation for future research to improve LLM performances in solving increasingly complex tasks.

%% file: contents/6-limitation.tex
\section*{Limitations}

While this study provides a comprehensive evaluation and analysis of LLMs’ rule-guided reasoning capabilities, there remain some limitations and numerous promising avenues for future research:

\paragraph{Automating Evaluation}
In this work, we only rely on GPT-4o to parse textual responses into structured JSON, facilitating downstream analyses of rule application. A logical next step would be to investigate the use of LLMs for fully automated reasoning evaluations, including the identification of intermediate errors. This direction aligns with the concept of ``LLM-as-a-judge'' \citep{zheng2023llmasajudge}, which, despite potential bias or inaccuracies \citep{xu2024pride}, offers a scalable alternative to labor-intensive human evaluation and could improve the reliability and granularity of assessment metrics.

\paragraph{Training with Rule-Guided Reasoning Data}
Supervised fine-tuning has proven effective in enhancing LLMs' performance on tasks requiring substantial domain knowledge \citep{jeong2024fine, fu2023specializing, wu2024akew}. While we did not pursue fine-tuning in this study—given the high cost of obtaining extensive rule-guided reasoning data and the limited generalizability to unseen domains—it remains a worthwhile direction. Investigating whether training with datasets from related domains, such as mathematical or logical reasoning tasks or code generation problems, can bolster LLMs’ rule-following performance is an open question worth exploring.

\paragraph{Sophisticated Rule Recall and Aggregation}
Our experiments reveal that LLMs frequently struggle with recalling and aggregating the correct rules, and that problem-wise recall strongly correlates with overall accuracy. Addressing these challenges may involve refining rule retrieval mechanisms or integrating structured reasoning frameworks. For instance, approaches that retrieve relevant information dynamically \citep{trivedi2023ircot,zhou2024trad} and convert rules into structured data \citep{pan2023logiclm,wang2024workingmemory} have shown promise in reasoning tasks. Building on these insights, a hybrid system that combines LLM-based reasoning with symbolic reasoners may enhance both the consistency and robustness of rule-guided reasoning in real-world scenarios.

%% file: contents/appendix.tex
\clearpage
\section{Terminology Explanation in NBA}\label{ap:NBA_terms}

We briefly explain the NBA terminologies mentioned in this paper as follows:

\paragraph{Salary Cap.} The Salary Cap of NBA is a rule that limits how much money each team can spend on player salaries. It is designed to keep teams on a level playing field financially, so wealthier teams cannot just purchase all the best players. The league sets the cap based on its overall revenue.

\paragraph{(Salary Cap) Exceptions.} The NBA uses a ``soft'' Salary Cap, meaning teams can exceed the limit using certain Exceptions. Following are some commonly used Exceptions:
\begin{itemize}
    \item \textbf{Mid-Level Exception (MLE)} allows teams to sign free players even if they are above the salary cap. There are three types of MLEs, i.e. Non-Taxpayer MLE, Taxpayer MLE, and MLE for Room Teams, applicable to teams in different salary situations.
    \item \textbf{Traded Player Exceptions (TPE)} is a tool that allows teams to make trades even if they are over the salary cap. When a team trades a player for less salary than it gives away (or for nothing), it creates a TPE, which is like a "credit" they can use later. If a team wants to acquire more salaries than it gives away in a trade, it can also use certain types of TPE to make such trade.
    \item \textbf{Veteran Free Agent Exception (Bird Rights)} in the NBA allow teams to re-sign their own players even if they are over the salary cap. Named after Larry Bird, this rule encourages teams to keep their star players. There are three types of Bird Rights, i.e. Bird Rights, Early Bird Rights, and Non-Bird Rights, applicable to players that play for the same team for different numbers of consecutive seasons.
\end{itemize}

\section{Data Collection and Annotation}

\subsection{Rule Collection}\label{ap:rules}

\paragraph{Airline.} We collect the policy for bag and optional fees from \emph{American Airlines}\footnote{\url{https://www.aa.com/i18n/customer-service/support/optional-service-fees.jsp}}. Specifically, the rules in the policy mainly include: 1) the allowance of carry-on luggage; 2) the base price for checking each luggage on different routes and in different cabin classes; 3) the additional fees for luggage overweight or oversize to varying degrees on different routes and in different cabin classes; 4) when calculating fees for overweight and oversize luggage for each piece, only the higher of the two should apply. Many rules (base price, overweight/oversize fees) in this domain are represented in tabular forms, and we regard one entire table as one rule.

\paragraph{NBA.} We collect the regulations for NBA transactions from \emph{2023 NBA Collective Bargaining Agreements}\footnote{\url{https://ak-static.cms.nba.com/wp-content/uploads/sites/4/2023/06/2023-NBA-Collective-Bargaining-Agreement.pdf}} (CBA) and excerpt from the \emph{NBA Constitution and By-Laws}\footnote{\url{https://ak-static-int.nba.com/wp-content/uploads/sites/3/2015/12/NBA-Constitution-and-By-Laws.pdf}} regarding the rules for trading first-round draft picks (i.e., the Stepien Rule). Since the complete CBA is too long (676 Pages PDF), we only aggregate the most commonly used rules such as the limits on salary and length of player contract, on team salary, and on player contract trade among teams. As applying rules of the same type but applicable under different conditions may result in completely different subsequent reasoning process, different from in airline domain, we depart such one paragraph including such similar rules into separate rules.

\paragraph{Tax.} We collect tax forms and relevant instructions from \emph{Internal Revenue Service} (IRS)\footnote{\url{https://www.irs.gov/forms-instructions}}. Starting from the most famous Form 1040 (U.S. Individual Income Tax Return) and its basic Schedules 1-3, we consider more complex settings commonly happen in real-life, including using itemized deductions (Schedule A), self-employment (Schedule C and Schedule SE), education expenses and/or credits (Form 8863), and child and/or other dependent credits (Schedule 8812). We treat each line in these forms and its instructions as one rule, and convert the forms into line numbers and text for each line as LLM input.

\subsection{NBA Data Annotation}\label{ap:annotation}

For NBA tasks, we first survey famous rules and transactions that have happened in the NBA in recent 30 years and decide the 54 rules used in our annotation. To balance the task difficulty and annotation difficulty, we further simplify the rules by unifying different types of team salary (defined in different rules and calculated in different ways) into one simple ``Team Salary''. The process of annotating one problem is described as follows:

\vspace{-4pt}\paragraph{Creating team and player situations.} Our annotators are first required to create diverse valid scenarios involving one or more teams and players, as the following ``team\_situations'' and ``players\_situations'', and provide the number of teams (``n\_teams'') and players (``n\_players'') involved. Each item in the ``team\_situations'' list indicates the current salary of the team and its available first-round draft picks, while each item in the``player\_situations'' list tells the player's information (i.e., draft year, age, and current (or last) contract). All players and teams are anonymized as Player (Team) A/B/C/... to avoid data leakage.

\vspace{-4pt}\paragraph{Writting transactions.} Next, our annotators write ``n\_operations'' sentences in the ``operations'' list, where each item corresponding to one team signing a player or several teams conducting a trade, and determine whether all these transactions are allowed according to the rules. The ``answer'' should be \textbf{True} if all transactions are allowed otherwise \textbf{False}. If ``answer'' is \textbf{False}, we ask our annotators to further provide ``illegal\_team'' and ``illegal\_operation'' as the specific team and transaction component that violates the rules.

\vspace{-4pt}\paragraph{Listing relevant rules.} Finally, our annotators are told to provide a list of ``relevant\_rules'' including all rules that they believe should be involved if humans need to consider the case comprehensively.

\begin{lstlisting}[language=python,style=pythonstyle,breakindent=0pt,title={The format of annotated NBA test problems.}]
{
    "n_teams": int = ...,
    "n_players": int = ...,
    "n_operations": int = ...,
    "team_situations": list[str] = [...],
    "player_situations": list[str] = [...],
    "operations": list[str] = [...],
    "answer": bool = ...,
    "illegal_operation": str = ...,
    "illegal_team": str = ...,
    "relevant_rules": list[str] = [...]
}
\end{lstlisting}

To ensure the quality of annotation, we provide each annotator with a detailed annotation document and training sessions, and ask our annotators to annotate a small subset of problems and give explanations for verification before annotation. Only if each instance in the verification subset is correct, the annotator will be invited for the formal annotation.

\section{Structured Rule Extraction in Each Scenario}\label{ap:parse}

As introduced in \se{subsec:exp-setting}, we utilize the structured output mode of GPT-4o \citep{openai2024gpt4o} to convert LLMs' textual output into structured data. Here we present the data structure we used in parsing.

\paragraph{Airline.} In airline domain we ask LLMs to parse the the list of checked luggage as well as provided basic information.

\begin{lstlisting}[language=python,style=pythonstyle,breakindent=0pt]
class BagCost(BaseModel):
    size: int
    weight: int
    base_check_fee: int
    oversize_fee: int
    overweight_fee: int
    total_fee: int
    
class PassengerClass(str, Enum):
    be = "Basic Economy"
    main = "Main Cabin"
    mp = "Main Plus"
    pe = "Premium Economy"
    business = "Business"
    first = "First"
    
class Response(BaseModel):
    passenger_class: str
    place_of_departure: str
    place_of_arrival: str
    ticket_price: int
    checked_bags: list[BagCost]
    total_cost: int
\end{lstlisting}

\paragraph{NBA.} In NBA domain we let the LLM parser decide whether each of the 54 rules is applied.

\begin{lstlisting}[language=python,style=pythonstyle,breakindent=0pt]
class RuleExtraction(BaseModel):
    # contract length
    contract_length_at_most_4_year_except_qualifying_veteran_free_agent_5_year: bool
    contract_length_at_most_2_year_bi_annual_exception: bool
    contract_length_at_most_4_year_non_taxpayer_mid_level_exception: bool
    contract_length_at_most_2_year_taxpayer_mid_level_exception: bool
    contract_length_at_most_3_year_mid_level_exception_for_room_team: bool
    contract_length_at_most_2_year_minimum_player_salary_exception: bool
    
    # basic rules
    salary_cap_no_exceed_without_exception: bool
    maximum_salary_for_player_less_than_7_year_service: bool
    maximum_salary_for_player_7_to_9_year_service: bool
    maximum_salary_for_player_10_or_more_year_service: bool
    higher_max_criterion_for_5th_year_eligible_player: bool
    salary_increase_and_decrease_ratio_except_qualiyfing_or_early_qualifying_veteran_free_agent: bool
    salary_increase_and_decrease_ratio_for_qualiyfing_or_early_qualifying_veteran_free_agent: bool
    
    # 38 year old provision
    defer_compensation_38_year_old: bool
    defer_compensation_qualifying_veteran_free_agent_38_year_old: bool
    
    # apron level as hard cap rules
    bi_annual_exception_hard_cap_first_apron_level: bool
    non_taxpayer_mid_level_exception_hard_cap_first_apron_level: bool
    sign_and_trade_hard_cap_first_apron_level: bool
    expanded_traded_player_exception_hard_cap_first_apron_level: bool
    aggregated_traded_player_exception_hard_cap_second_apron_level: bool
    cash_in_trade_hard_cap_second_apron_level: bool
    sign_and_trade_assigner_traded_player_exception_hard_cap_second_apron_level: bool
    taxpayer_mid_level_exception_hard_cap_second_apron_level: bool
    traded_player_exception_250k_reduced_first_apron_level: bool
    
    # exceptions
    # bird rights
    qualifying_veteran_free_agent_exception: bool
    early_qualifying_veteran_free_agent_exception: bool
    non_qualifying_veteran_free_agent_exception: bool
    salary_space_consumption_qualifying_veteran_free_agent: bool
    salary_space_consumption_early_qualifying_veteran_free_agent: bool
    salary_space_consumption_non_qualifying_veteran_free_agent: bool
    salary_space_consumption_standard_traded_player_exception: bool
    
    # bi-annual exception
    bi_annual_exception: bool
    
    # mid level exceptions
    non_taxpayer_mid_level_exception: bool
    taxpayer_mid_level_exception: bool
    mid_level_exception_for_room_team: bool
    minimum_player_salary_exception: bool
    
    # traded player exceptions
    standard_traded_player_exception: bool
    aggregated_standard_traded_player_exception: bool
    expanded_traded_player_exception: bool
    traded_player_exception_for_room_team: bool
    traded_player_exception_only_one_minimum_traded_player_under_conditions: bool
    
    # trade rules
    pay_or_receive_cash_maximum_in_a_year: bool
    rookie_or_two_way_contract_cannot_be_traded_within_30_days: bool
    free_agent_sign_contract_cannot_be_traded_within_3_month_or_before_dec_15: bool
    qualifying_or_early_qualifying_free_agent_sign_contract_cannot_be_traded_within_3_month_or_before_jan_15: bool
    
    # sign-and-trade rules
    sign_and_trade_3_to_4_year: bool
    sign_and_trade_not_with_mid_level_exception: bool
    sign_and_trade_no_higher_than_25_percent_for_higher_max_5th_year_eligible_player: bool
    sign_and_trade_assignee_team_has_room: bool
    sign_and_trade_qualifying_free_agent_half_salary_for_traded_player_exception: bool
    
    # restricted free agent rules (Arenas provision)
    offer_sheet_for_1_or_2_year_service_player_no_more_than_mid_level_in_first_2_year: bool
    offer_sheet_for_1_or_2_year_service_player_3rd_year_maximum_if_first_2_year_maximum: bool
    offer_sheet_for_1_or_2_year_service_player_4th_year_maximum_if_3_year: bool
    offer_sheet_for_1_or_2_year_service_player_average_salary_more_than_2_year: bool
    
    # first-round draft pick trade rules
    stepien_rule_no_sell_or_no_consecutive_first_round_draft_pick_trade: bool
\end{lstlisting}

\paragraph{Tax.} In tax domain we just list each line in Form 1040 and its Schedules 1-3 for parsing.
\begin{lstlisting}[language=python,style=pythonstyle,breakindent=0pt]
class Form1040(BaseModel):
    name: str = Field(description="Name of taxpayer")
    age: int = Field(description="Age of taxpayer")
    spouse_age: int = Field(description="Age of taxpayer's spouse")
    filing_status: FilingStatus = Field(description="Filing status of taxpayer")
    blind: bool = Field(description="Taxpayer is blind")
    spouse_blind: bool = Field(description="Taxpayer's spouse is blind")
    itemized: bool = Field(description="Taxpayer uses itemized deductions")
    num_qualifying_children: int = Field(description="Number of qualifying children")
    num_other_dependents: int = Field(description="Number of other dependents")
    wage_tip_compensation: float = Field(description="Form 1040 Line 1a")
    household_employee_wage: float = Field(description="Form 1040 Line 1b")
    unreported_tip: float = Field(description="Form 1040 Line 1c")
    nontaxable_combat_pay: float = Field(description="Form 1040 Line 1d")
    wage_tip_compensation_total: float = Field(description="Form 1040 Line 1z")
    tax_exempt_interest: float = Field(description="Form 1040 Line 2a")
    taxable_interest: float = Field(description="Form 1040 Line 2b")
    qualified_dividends: float = Field(description="Form 1040 Line 3a")
    ordinary_dividends: float = Field(description="Form 1040 Line 3b")
    ira_distributions: float = Field(description="Form 1040 Line 4a")
    taxable_ira_distributions: float = Field(description="Form 1040 Line 4b")
    all_pensions: float = Field(description="Form 1040 Line 5a")
    taxable_pensions: float = Field(description="Form 1040 Line 5b")
    social_security_benefits: float = Field(description="Form 1040 Line 6a")
    taxable_social_security_benefits: float = Field(description="Form 1040 Line 6b")
    capital_gain_or_loss: float = Field(description="Form 1040 Line 7")
    additional_income: float = Field(description="Form 1040 Line 8")
    total_income: float = Field(description="Form 1040 Line 9")
    total_adjustments: float = Field(description="Form 1040 Line 10")
    adjusted_gross_income: float = Field(description="Form 1040 Line 11")
    standard_or_itemized_deductions: float = Field(description="Form 1040 Line 12")
    qualified_business_income: float = Field(description="Form 1040 Line 13")
    total_deductions: float = Field(description="Form 1040 Line 14")
    computed_taxable_income: float = Field(description="Form 1040 Line 15")
    taxes: float = Field(description="Form 1040 Line 16")
    copy_schedule_2_line_3: float = Field(description="Form 1040 Line 17")
    f1040_line_18: float = Field(description="Form 1040 Line 18")
    ctc_or_other_dependent_credit: float = Field(description="Form 1040 Line 19")
    copy_schedule_3_line_8: float = Field(description="Form 1040 Line 20")
    accumulated_credits: float = Field(description="Form 1040 Line 21")
    taxes_after_credits: float = Field(description="Form 1040 Line 22")
    other_taxes: float = Field(description="Form 1040 Line 23")
    total_tax: float = Field(description="Form 1040 Line 24")
    federal_income_tax_withheld: float = Field(description="Form 1040 Line 25")
    earned_income_credit: float = Field(description="Form 1040 Line 27")
    additional_child_tax_credit: float = Field(description="Form 1040 Line 28")
    american_opportunity_credit: float = Field(description="Form 1040 Line 29")
    copy_schedule_3_line_15: float = Field(description="Form 1040 Line 31")
    total_other_payments_and_refundable_credits: float = Field(description="Form 1040 Line 32")
    total_payments: float = Field(description="Form 1040 Line 33")
    amount_owed_or_overpaid: float = Field(description="Form 1040 Line 37 (negative if overpaid)")
    taxable_state_refunds: float = Field(description="Schedule 1 Line 1")
    alimony_income: float = Field(description="Schedule 1 Line 2a")
    sale_of_business: float = Field(description="Schedule 1 Line 4")
    rental_real_estate_sch1: float = Field(description="Schedule 1 Line 5")
    farm_income: float = Field(description="Schedule 1 Line 6")
    unemployment_compensation: float = Field(description="Schedule 1 Line 7")
    other_income: float = Field(description="Schedule 1 Line 8")
    educator_expenses: float = Field(description="Schedule 1 Line 11")
    hsa_deduction: float = Field(description="Schedule 1 Line 13")
    self_employment_deductible: float = Field(description="Schedule 1 Line 15")
    ira_deduction: float = Field(description="Schedule 1 Line 20")
    student_loan_interest_deduction: float = Field(description="Schedule 1 Line 21")
    other_adjustments: float = Field(description="Schedule 1 Line 24")
    amt_f6251: float = Field(description="Schedule 2 Line 1")
    credit_repayment: float = Field(description="Schedule 2 Line 2")
    schedule_2_total_taxes: float = Field(description="Schedule 2 Line 3 (= Line 1 + Line 2)")
    self_employment_tax: float = Field(description="Schedule 2 Line 4")
    other_additional_taxes: float = Field(description="Schedule 2 Line 17")
    schedule_2_total_other_taxes: float = Field(description="Schedule 2 Line 21 (= Line 4 + Line 17)")
    foreign_tax_credit: float = Field(description="Schedule 3 Line 1")
    dependent_care: float = Field(description="Schedule 3 Line 2")
    computed_education_credits: float = Field(description="Schedule 3 Line 3")
    retirement_savings: float = Field(description="Schedule 3 Line 4")
    elderly_disabled_credits: float = Field(description="Schedule 3 Line 6d")
    plug_in_motor_vehicle: float = Field(description="Schedule 3 Line 6i")
    alt_motor_vehicle: float = Field(description="Schedule 3 Line 6j")
    schedule_3_line_8: float = Field(description="Schedule 3 Line 8")
    medical_dental_expenses: Optional[float] = Field(description="Schedule A Line 1 (if itemized)")
    state_local_income_or_sales_tax: Optional[float] = Field(description="Schedule A Line 5a (if itemized)")
    state_local_real_estate_tax: Optional[float] = Field(description="Schedule A Line 5b (if itemized)")
    state_local_personal_property_tax: Optional[float] = Field(description="Schedule A Line 5c (if itemized)")
    other_taxes_paid: Optional[float] = Field(description="Schedule A Line 6 (if itemized)")
    home_mortgage_interest_and_points: Optional[float] = Field(description="Schedule A Line 8a (if itemized)")
    home_mortgage_interest_unreported: Optional[float] = Field(description="Schedule A Line 8b")
    home_mortgage_points_unreported: Optional[float] = Field(description="Schedule A Line 8c (if itemized)")
    investment_interest: Optional[float] = Field(description="Schedule A Line 9 (if itemized)")
    charity_cash: Optional[float] = Field(description="Schedule A Line 11 (if itemized)")
    charity_non_cash: Optional[float] = Field(description="Schedule A Line 12 (if itemized)")
    casualty_and_theft_loss: Optional[float] = Field(description="Schedule A Line 15 (if itemized)")
    other_itemized_deductions: Optional[float] = Field(description="Schedule A Line 16 (if itemized)")
    gross_receipts: Optional[float] = Field(description="Schedule C Line 1 (if self-employed)")
    returns_and_allowances: Optional[float] = Field(description="Schedule C Line 2 (if self-employed)")
    cost_of_goods_sold: Optional[float] = Field(description="Schedule C Line 4 (if self-employed)")
    other_inc_sched_c: Optional[float] = Field(description="Schedule C Line 6 (if self-employed)")
    total_expenses: Optional[float] = Field(description="Schedule C Line 28 (if self-employed)")
    expenses_of_home: Optional[float] = Field(description="Schedule C Line 30 (if self-employed)")
    net_profit: Optional[float] = Field(description="Schedule C Line 31 (if self-employed)")
    total_social_security_wages: Optional[float] = Field(description="Schedule SE Line 8 (if self-employed)")
    student_list: Optional[list[Student]] = Field(description="List of students with education expenses")
\end{lstlisting}

\section{More Experiment Results and Analysis}\label{ap:exp}

\subsection{Rule-Wise Statistics}\label{ap:rule-wise-stats}

\begin{figure}[htbp]
    \centering
    \begin{subfigure}[b]{0.48\textwidth}
        \includegraphics[width=\linewidth]{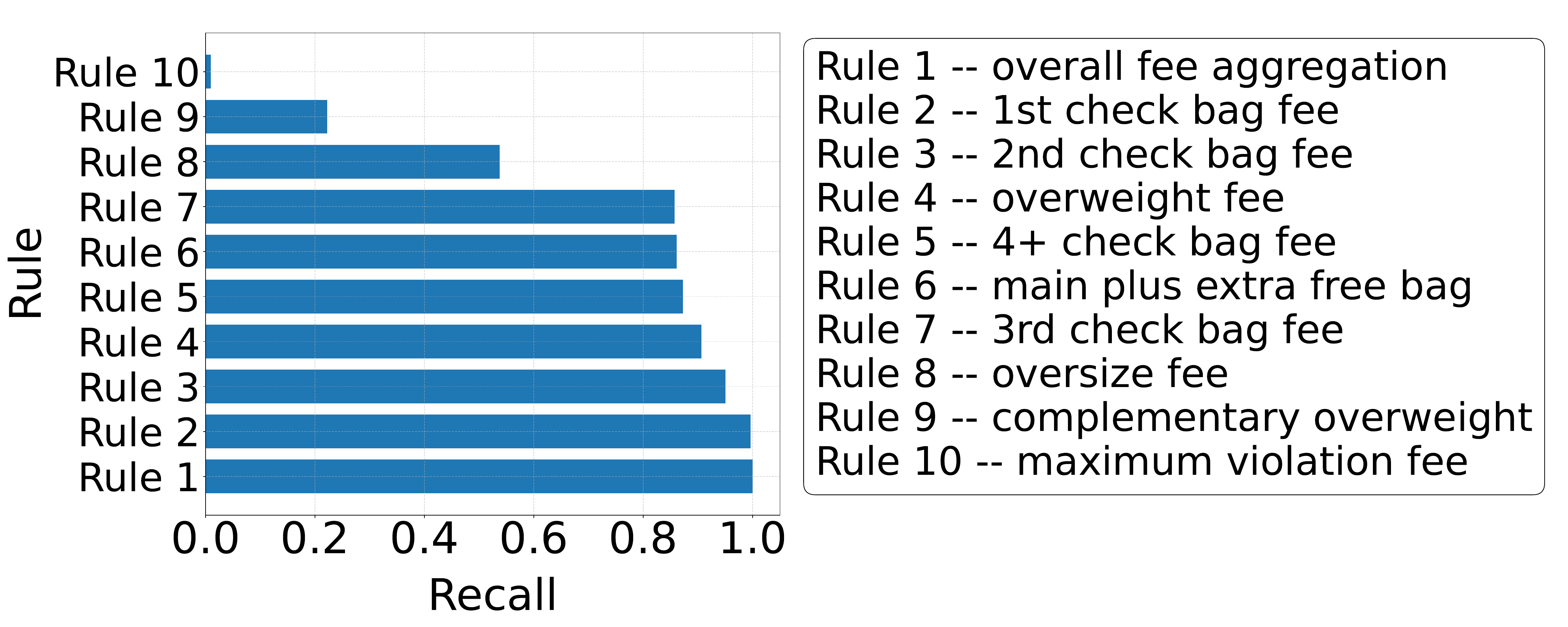}
        \caption{Recall}
        \label{fig:rule_wise_recall_airline_ap}
    \end{subfigure}~~
    \begin{subfigure}[b]{0.48\textwidth}
        \includegraphics[width=\linewidth]{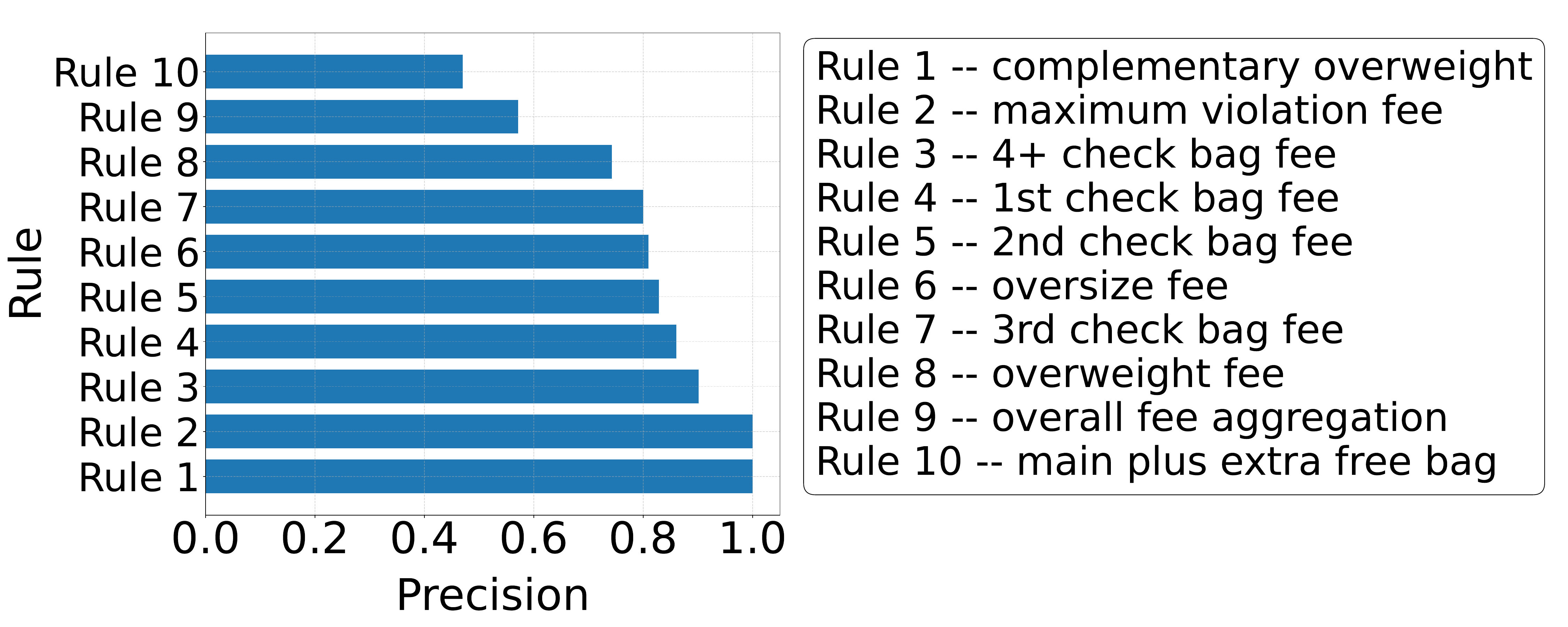}
        \caption{Correctness}
        \label{fig:rule_wise_correctness_airline_ap}
    \end{subfigure}
    \caption{Rule-wise metrics of rules in airline domain.}
    \label{fig:rule_wise_airline_ap}
\end{figure}

\begin{figure}[htbp]
    \centering
    \includegraphics[width=\linewidth]{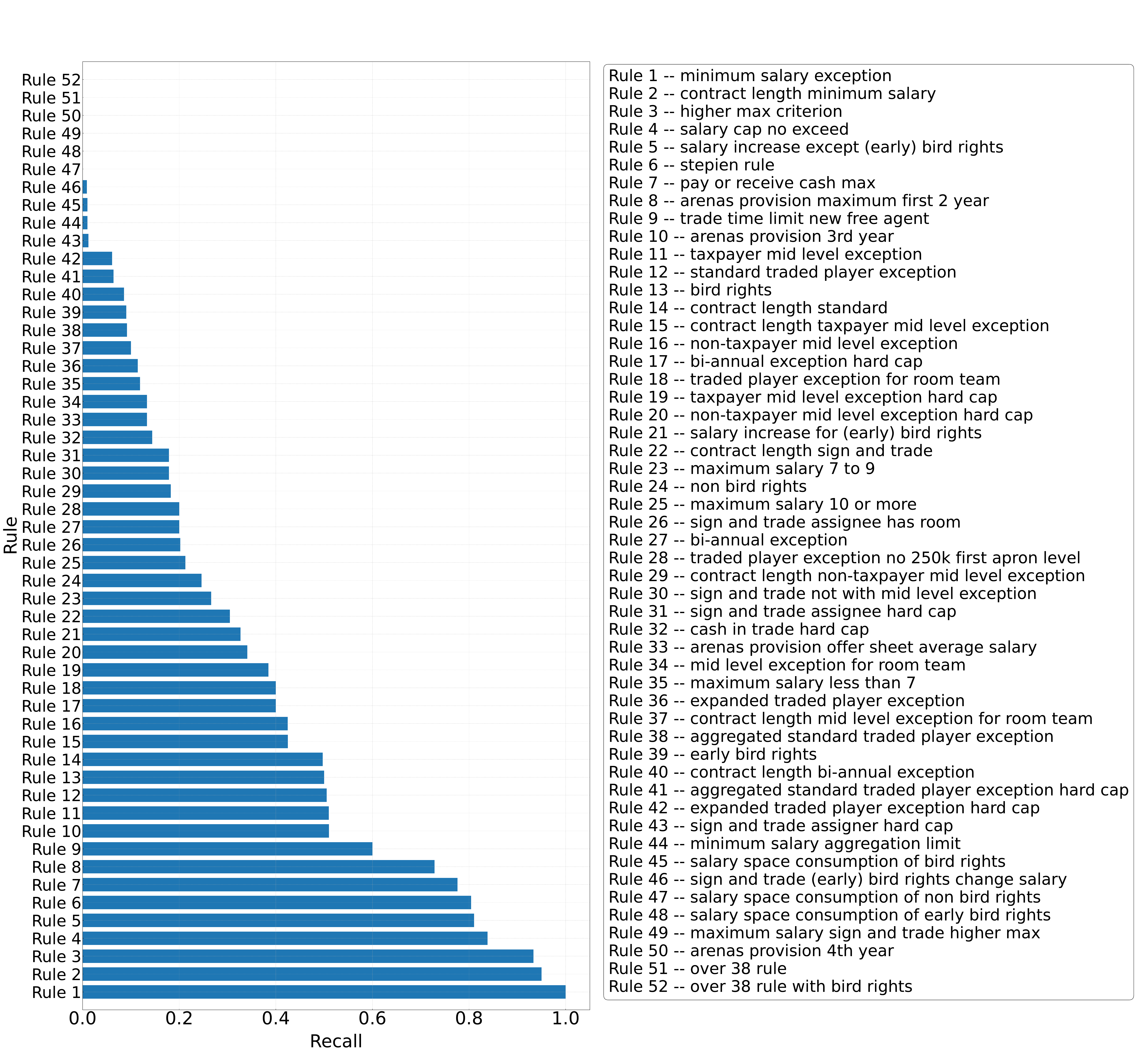}
    \caption*{(a) Recall}
    \label{fig:rule_wise_recall_nba}
\end{figure}

\begin{figure}[htbp]
    \begin{subfigure}[b]{\textwidth}
    \centering
        \includegraphics[width=\linewidth]{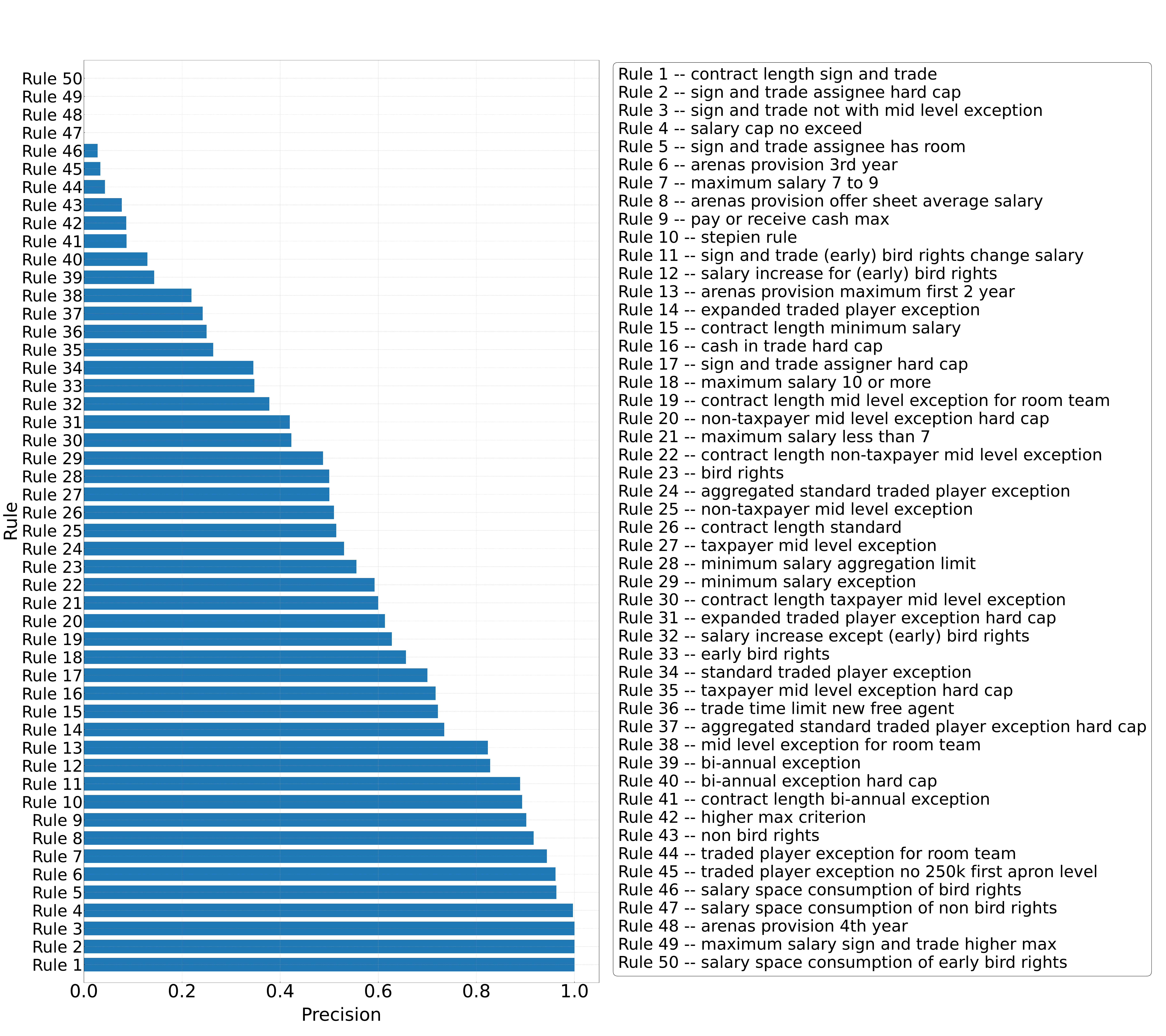}
        \caption*{(b) Precision}
        \label{fig:rule_wise_precision_nba}
    \end{subfigure}
    \caption{Rule-wise metrics of rules in NBA domain.}
    \label{fig:rule_wise_nba}
\end{figure}

\begin{figure}[htbp]
    \centering
    \begin{subfigure}[b]{0.48\textwidth}
        \includegraphics[width=\linewidth]{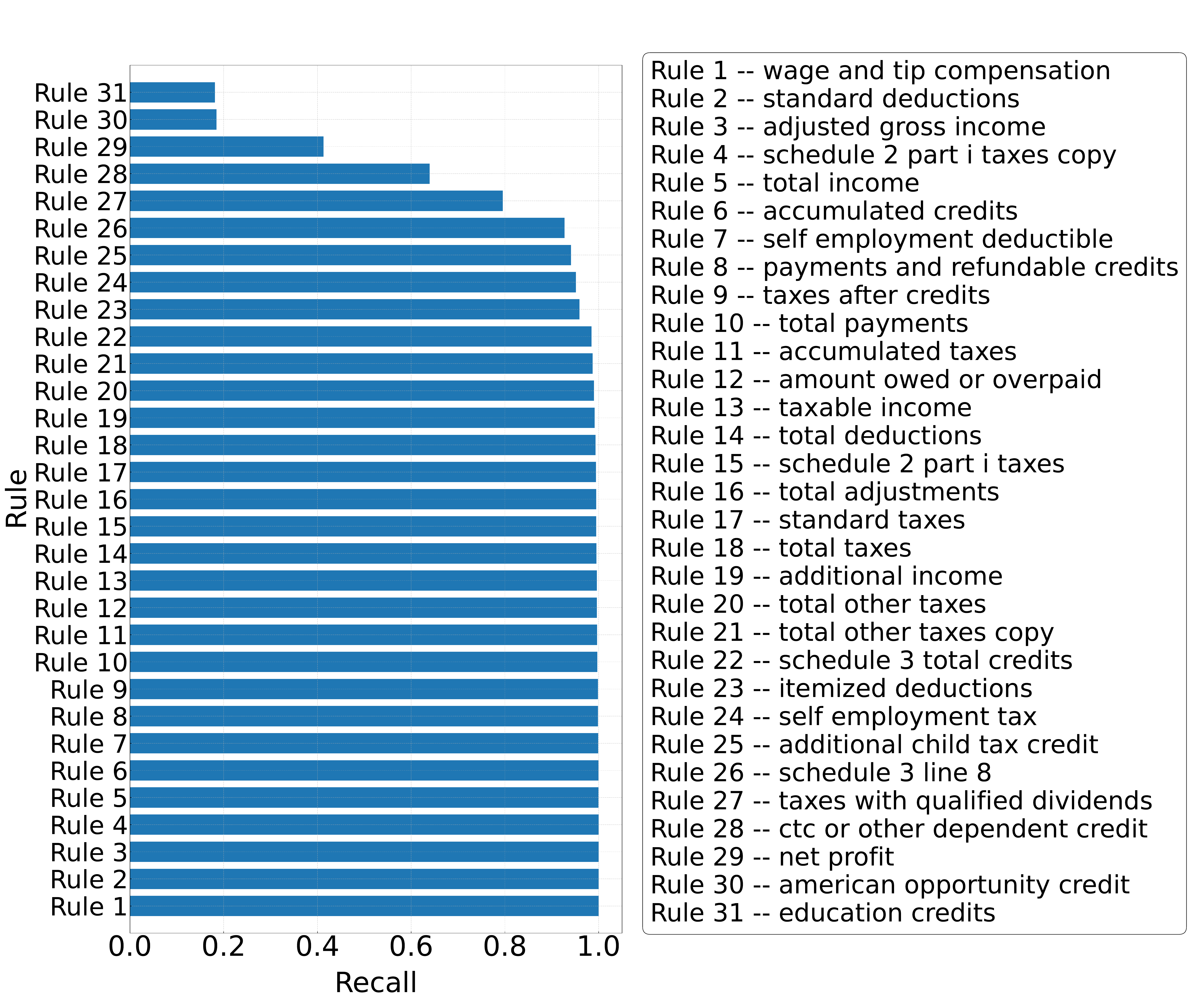}
        \caption{Recall}
        \label{fig:rule_wise_recall_tax}
    \end{subfigure}~~
    \begin{subfigure}[b]{0.48\textwidth}
        \includegraphics[width=\linewidth]{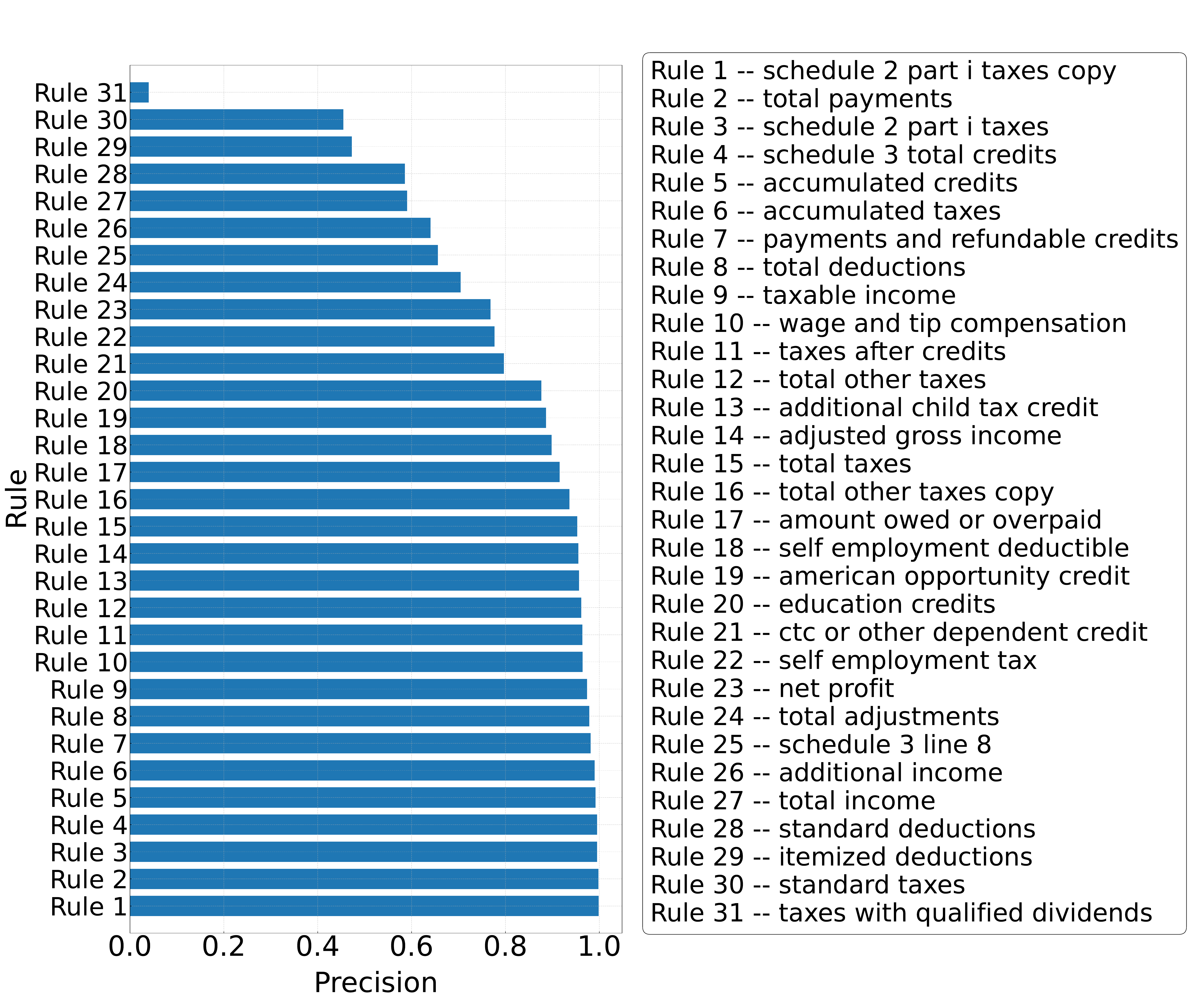}
        \caption{Correctness}
        \label{fig:rule_wise_correctness_tax}
    \end{subfigure}
    \caption{Rule-wise metrics of rules in tax domain.}
    \label{fig:rule_wise_tax}
\end{figure}

We visualize the rule-wise \emph{recall}, \emph{precision}, and \emph{correctness} in Figure \ref{fig:rule_wise_airline_ap}-\ref{fig:rule_wise_tax}. Since \emph{precision} is always $1.0$ in airline and tax domains, we skip these two charts.


\subsection{What Impacts Rule Following?}\label{ap:rule_following_factor}

In this section, we investigate the factors influencing LLM performance, as measured by ${\rm Acc}(t)$. We begin by examining the correlation between ${\rm Acc}(t)$ and other key metrics, including ${\rm P}(t)$, ${\rm AC}(t)$, and ${\rm R}(t)$. We then consider the effects of in-context examples, different rule representations, and the presence of distractors.

\subsubsection{Correlation Between Accuracy and Other Metrics}\label{ap:corr-between-metrics}
\begin{figure*}[!ht]
  \centering
  \begin{subfigure}[b]{0.32\textwidth}
    \centering
    \includegraphics[width=\linewidth]{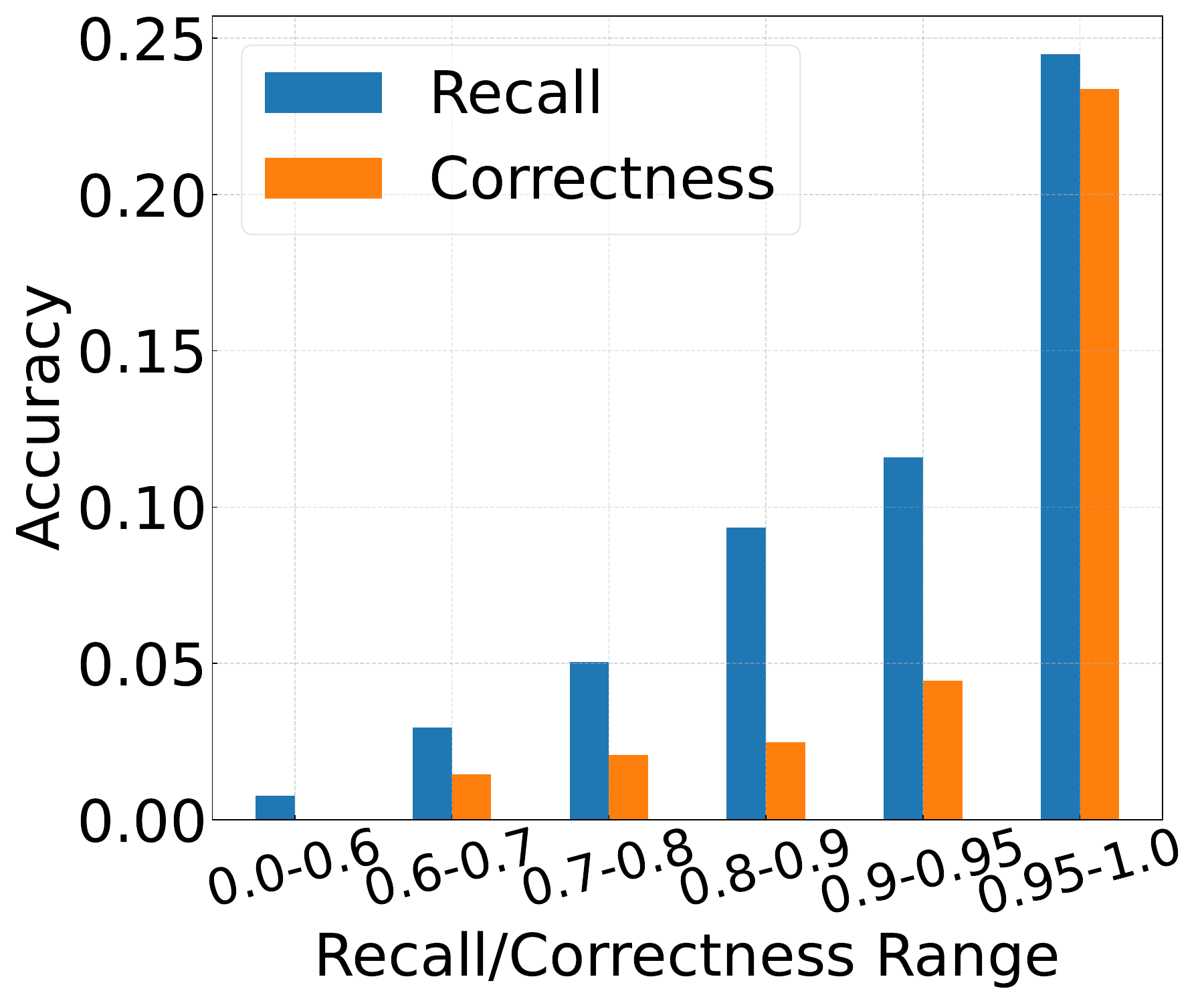}
    \caption{Airline}
    \label{fig:corr_airline}
  \end{subfigure}~~
  \begin{subfigure}[b]{0.32\textwidth}
    \centering
    \includegraphics[width=\linewidth]{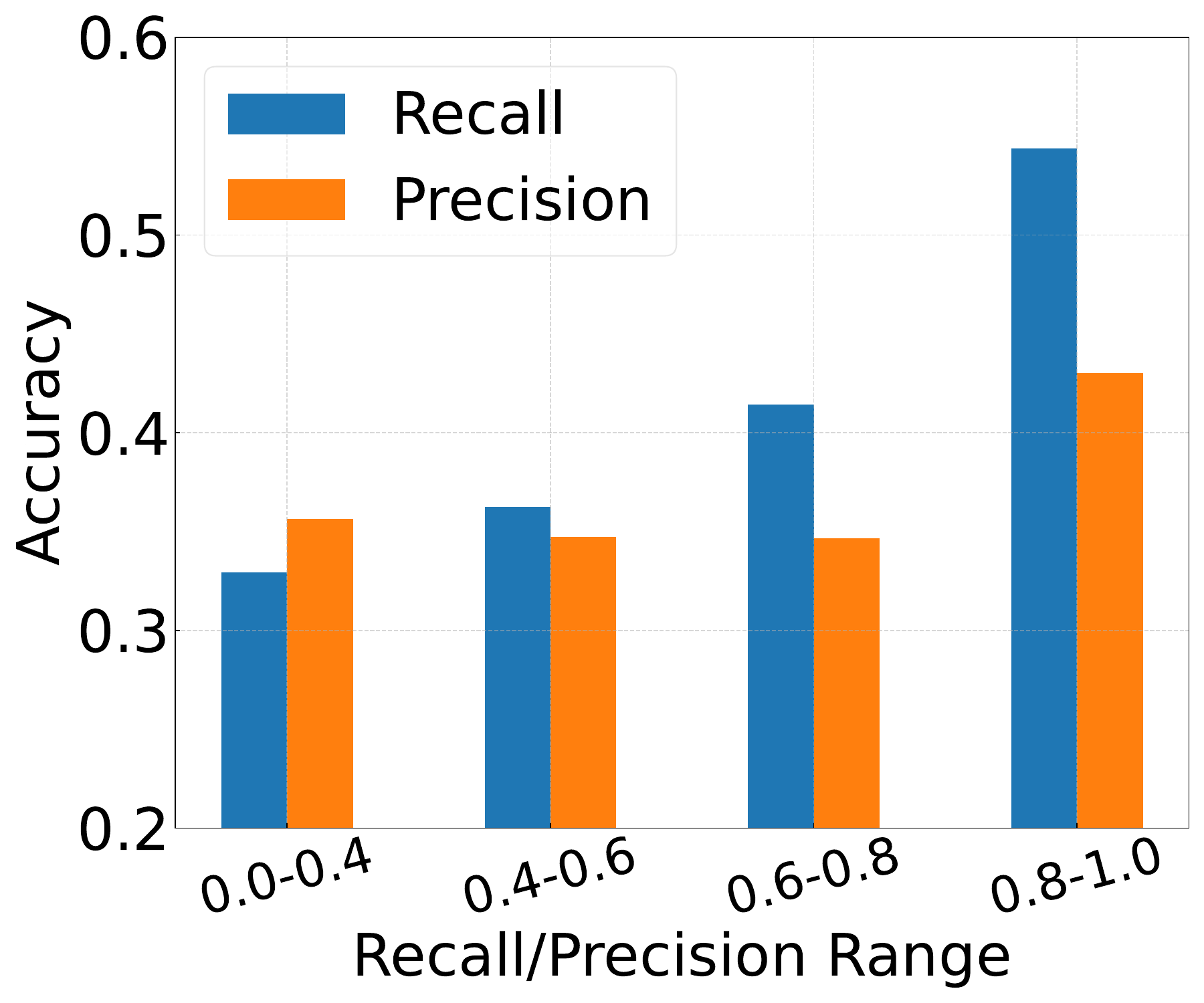}
    \caption{NBA}
    \label{fig:corr_nba}
  \end{subfigure}~~
  \begin{subfigure}[b]{0.32\textwidth}
    \centering
    \includegraphics[width=\linewidth]{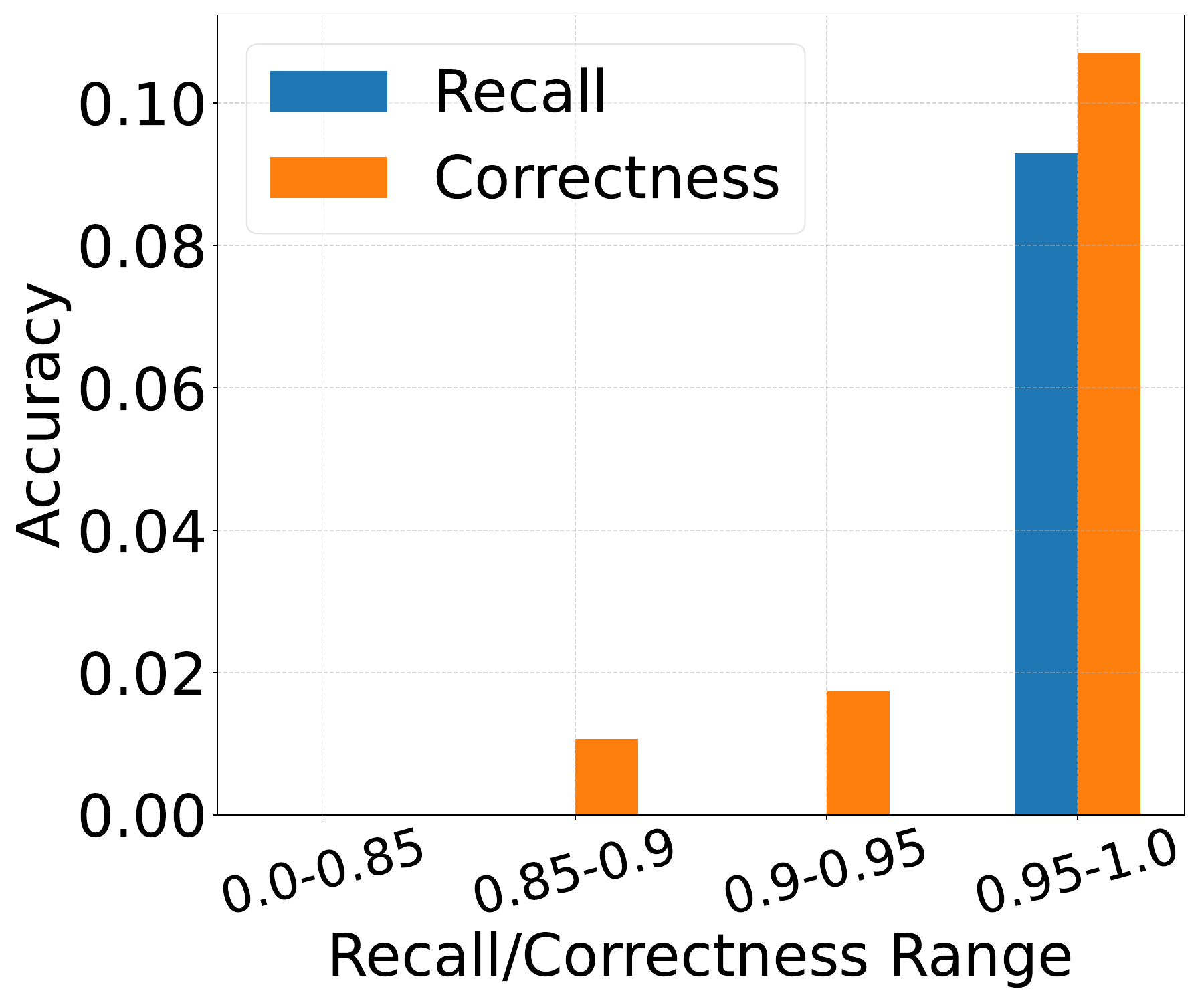}
    \caption{Tax}
    \label{fig:corr_tax}
  \end{subfigure}
  \vspace{-4pt}
  \caption{Correlation between problem-wise metrics and accuracy. The correlation is the most obvious and almost linear between ${\rm R}(t)$ and ${\rm Acc}(t)$, while highly non-linear or unclear between other two metrics and ${\rm Acc}(t)$.}
  \label{fig:corr}
\end{figure*}

To understand which factors most directly affect ${\rm Acc}(t)$, we visualize its correlation with other metrics in Figure \ref{fig:corr} across all three domains on datapoints from all difficulty levels. From Figure \ref{fig:corr_airline} and Figure \ref{fig:corr_nba}, we observe an almost linear relationship between ${\rm R}(t)$ and ${\rm Acc}(t)$. Notice that in the tax domain (Figure \ref{fig:corr_tax}), a recall lower than $0.95$ immediately results in zero accuracy.

In contrast, the correlation between ${\rm AC}(t)$ and ${\rm Acc}(t)$ is highly non-linear, as seen in Figure \ref{fig:corr_airline} and Figure \ref{fig:corr_tax}. In many cases, a single computational error in rule application (thus reducing ${\rm AC}(t)$) is sufficient to produce an incorrect final answer, indicating that only near-perfect ${\rm AC}(t)$ leads to significant ${\rm Acc}(t)$ improvements. For the NBA domain, we also compare ${\rm P}(t)$ and ${\rm Acc}(t)$; since ${\rm P}(t)$ is always 100\% for the airline and tax domains, these correlations are not meaningful there. We find no clear relationship between ${\rm P}(t)$ and ${\rm Acc}(t)$ for the NBA problems (Figure \ref{fig:corr_nba}).

\subsubsection{Do In-Context Examples Help?}\label{ap:in-context-example}
Table~\ref{tab:main_table} presents the results with or without a level-1 1-shot example. LLMs generally provide better performances given 1-shot example on airline, tax, and (easy) NBA problems. Many studies have shown the benefit of in-context learning \citep{dong2022survey, wei2023larger, zhang2023ideal}, which conforms with our observation that ${\rm Acc}(t)$ gets higher in the 1-shot setting. This performance boost comes from both the enhancement of ${\rm AC}(t)$ as well as a better understanding of the reasoning process, indicated by higher ${\rm R}(t)$.

However, when tackling more challenging NBA problems (Levels 2 and 3), providing an example increases ${\rm P}(t)$ and ${\rm R}(t)$ but leads to a counterintuitive decrease in overall ${\rm Acc}(t)$. This improvement in precision and recall primarily arises from the non-essential rules included in the in-context example, such as the ``Over 38 rule'' and ``Salary consumption of veteran free agent''. We compute the ${\rm R}(r)$ and ${\rm P}(r)$ for these two rules as in Table~\ref{tab:rule_wise_1shot}.

\begin{table}[!ht]
    \centering
    \resizebox{0.5\columnwidth}{!}{
    \begin{tabular}{lccc}
    \toprule
        \multicolumn{1}{c}{\textbf{Rule}} & \textbf{Setting} & ${\rm R}(r)$ & ${\rm P}(r)$ \\
    \midrule
        \multirow{2}{*}{Over 38 rule} & 0-shot & 0.00 & N/A \\
        & 1-shot & 0.35 & 0.76 \\
    \hdashline
        \multirow{2}{*}{Salary consumption} & 0-shot & 0.00 & N/A \\
        & 1-shot & 0.23 & 0.20 \\
    \bottomrule
    \end{tabular}
    }
    \caption{0-shot and 1-shot rule-wise comparison.}
    \label{tab:rule_wise_1shot}
\end{table}

Notably, while the ${\rm R}(r)$ for both rules improves, ${\rm P}(r)$ for rule ``Salary consumption'' is much lower. This shows that thought the in-context example does remind LLMs to apply rules that they might overlook, some rules like ``Salary consumption'' can be too hard for LLMs to understand even taught by an expert example, and thus LLMs do not understand what scenarios are suitable for such rules to apply.
In addition, we find the performance on the remaining rules remains mostly unchanged. The exact cause of the performance decline in accuracy is difficult to pinpoint as our annotation on NBA does not contain detailed intermediate reasoning annotations. However, prior work \citep{fan2023nphardeval} suggests that if the in-context example is ``easier'' than the target problem, the example can inadvertently degrade performance---a plausible explanation for why accuracy drops even as precision and recall improve.

\subsubsection{Does Rule Representation Matter?}\label{ap:rule-representation}

In the airline and tax domains, some rules are represented as Markdown tables. To test whether representation format affects performance, we convert these tabular rules into textual ``if-then'' statements. Table~\ref{tab:rule_repr} shows that converting tabular rules into text improves ${\rm R}(r)$, but has little impact on other metrics, including ${\rm Acc}(t)$.

\begin{table}[!ht]
    \centering
    \resizebox{12cm}{!}{
    \begin{tabular}{lccccccc}
        \toprule
        \multirow{2}{*}{\textbf{Models}} & \multirow{2}{*}{\textbf{Setting}} & \multicolumn{3}{c}{\textbf{Airline}} & \multicolumn{3}{c}{\textbf{Tax}} \\
        \cmidrule(lr){3-5} \cmidrule(lr){6-8} & & ${\rm AC}(t)$ & ${\rm R}(t)$ & ${\rm Acc}(t)$ & ${\rm AC}(t)$ & ${\rm R}(t)$ & ${\rm Acc}(t)$\\
        \midrule
        \multirow{2}{*}{Llama 70B} & Table & 0.764 & 0.558 & 0.01 & 0.834 & 0.989 & 0.01 \\
         & Text & 0.764 & 0.582 & 0.01 & 0.814 & 0.991 & 0.00 \\
        \hdashline
        \multirow{2}{*}{Qwen 72B} & Table & 0.636 & 0.586 & 0.01 & 0.888 & 0.998 & 0.10 \\
         & Text & 0.748 & 0.633 & 0.02 & 0.859 & 0.996 & 0.01 \\
        \hdashline
        \multirow{2}{*}{Llama 405B} & Table & 0.854 & 0.604 & 0.03 & 0.923 & 0.999 & 0.16 \\
         & Text & 0.835 & 0.587 & 0.07 & 0.919 & 0.998 & 0.05 \\
        \hdashline
        \multirow{2}{*}{Claude-3.5} & Table & 0.930 & 0.702 & 0.04 & 0.964 & 1.000 & 0.32 \\
         & Text & 0.937 & 0.705 & 0.06 & 0.971 & 1.000 & 0.33 \\
        \hdashline
        \multirow{2}{*}{GPT-4o} & Table & 0.862 & 0.616 & 0.02 & 0.965 & 1.000 & 0.42 \\
         & Text & 0.864 & 0.669 & 0.03 & 0.960 & 1.000 & 0.33 \\
        \bottomrule
    \end{tabular}
    }
    \caption{Results of different LLMs given different rule representations.}
    \label{tab:rule_repr}
\end{table}

\subsubsection{Do Distractive Rules Matter?}\label{ap:distractive-rules}

An essential aspect of rule-following involves identifying which rules are relevant to the current problem. In our experiments, all domain-specific rules are provided in the prompt, leaving it to the LLMs to determine which ones should be applied. To assess the extent to which irrelevant rules detrimentally affect performance, we focus on the tax scenario. In this domain, we can introduce additional tax forms that contain only zero values, effectively rendering any corresponding rules irrelevant. Despite these rules being unnecessary, their mere presence may mislead LLMs into treating them as important.

To isolate the effect of these distractive rules from the influence of increased context length, we also create a ``Placeholder'' setting. In this setting, we replace the distractive rules with an equivalent amount of meaningless tokens that do not correspond to any rules. By comparing performance under these two conditions, we can distinguish between the impact of irrelevant rules and the general challenge posed by a longer input.

\begin{figure*}[!ht]
  \centering
  \begin{subfigure}[b]{0.32\textwidth}
    \centering
    \includegraphics[width=\linewidth]{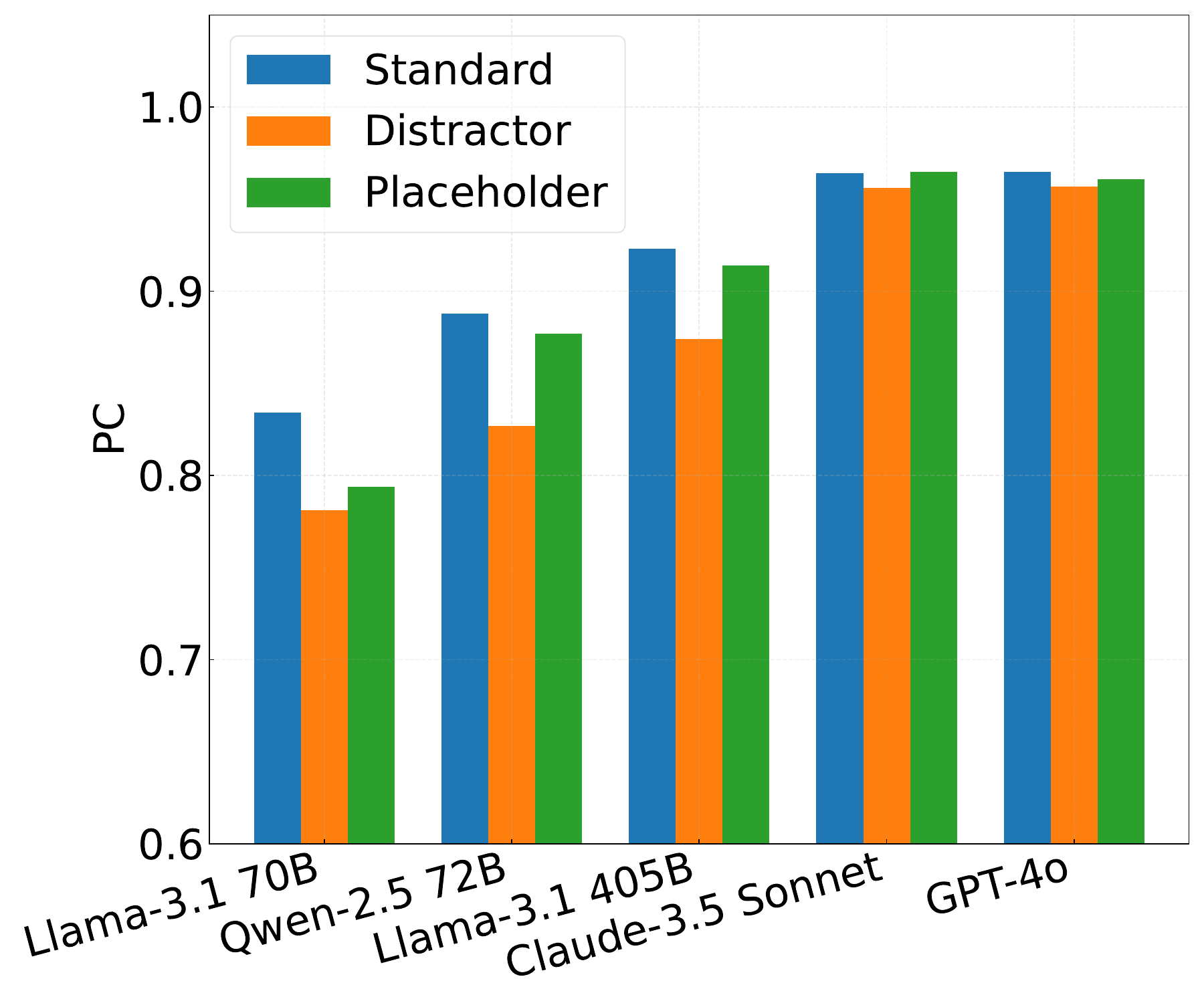}
    \caption{Correctness}
    \label{fig:context_pc}
  \end{subfigure}~~
  \begin{subfigure}[b]{0.32\textwidth}
    \centering
    \includegraphics[width=\linewidth]{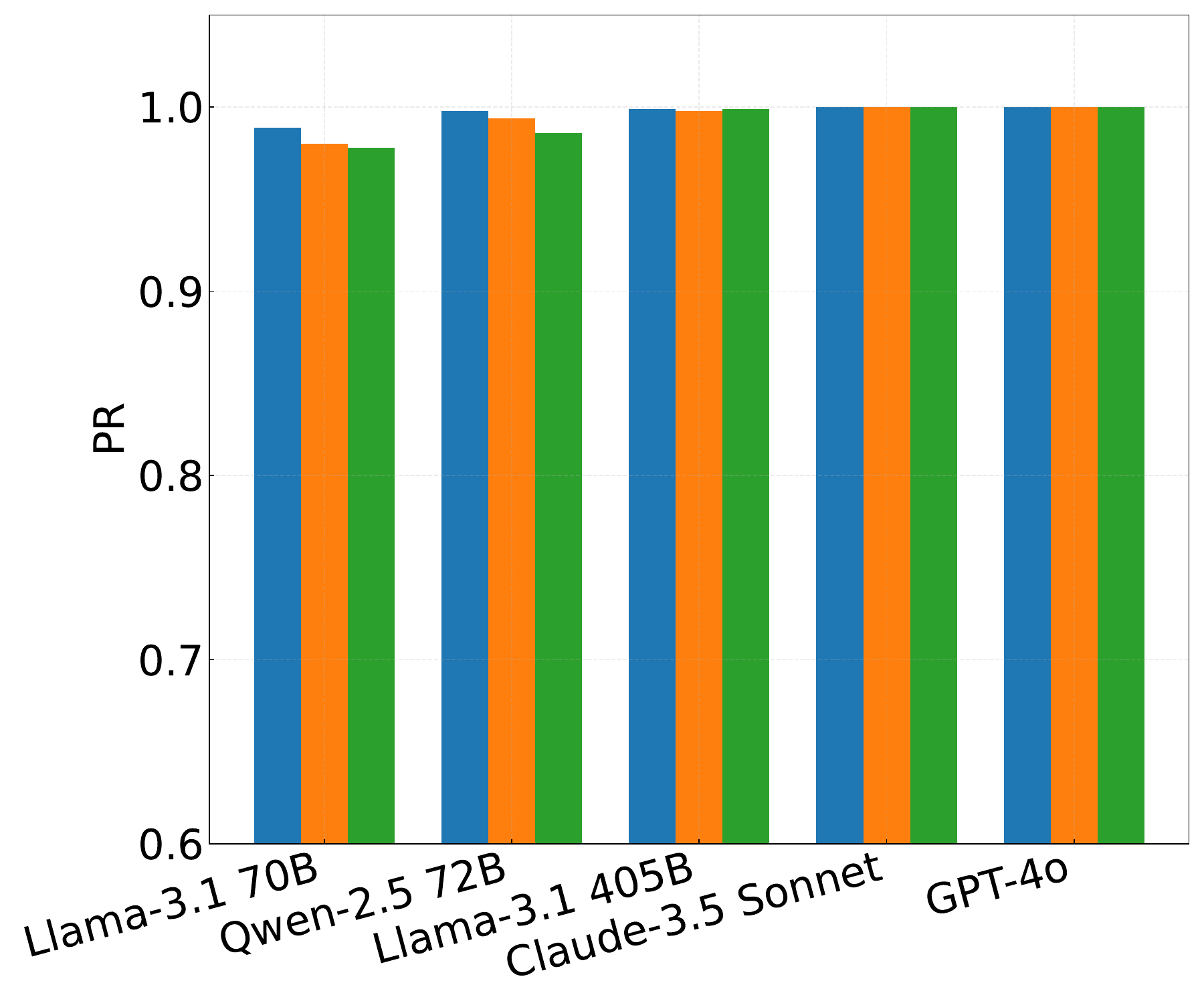}
    \caption{Recall}
    \label{fig:context_pr}
  \end{subfigure}~~
  \begin{subfigure}[b]{0.32\textwidth}
    \centering
    \includegraphics[width=\linewidth]{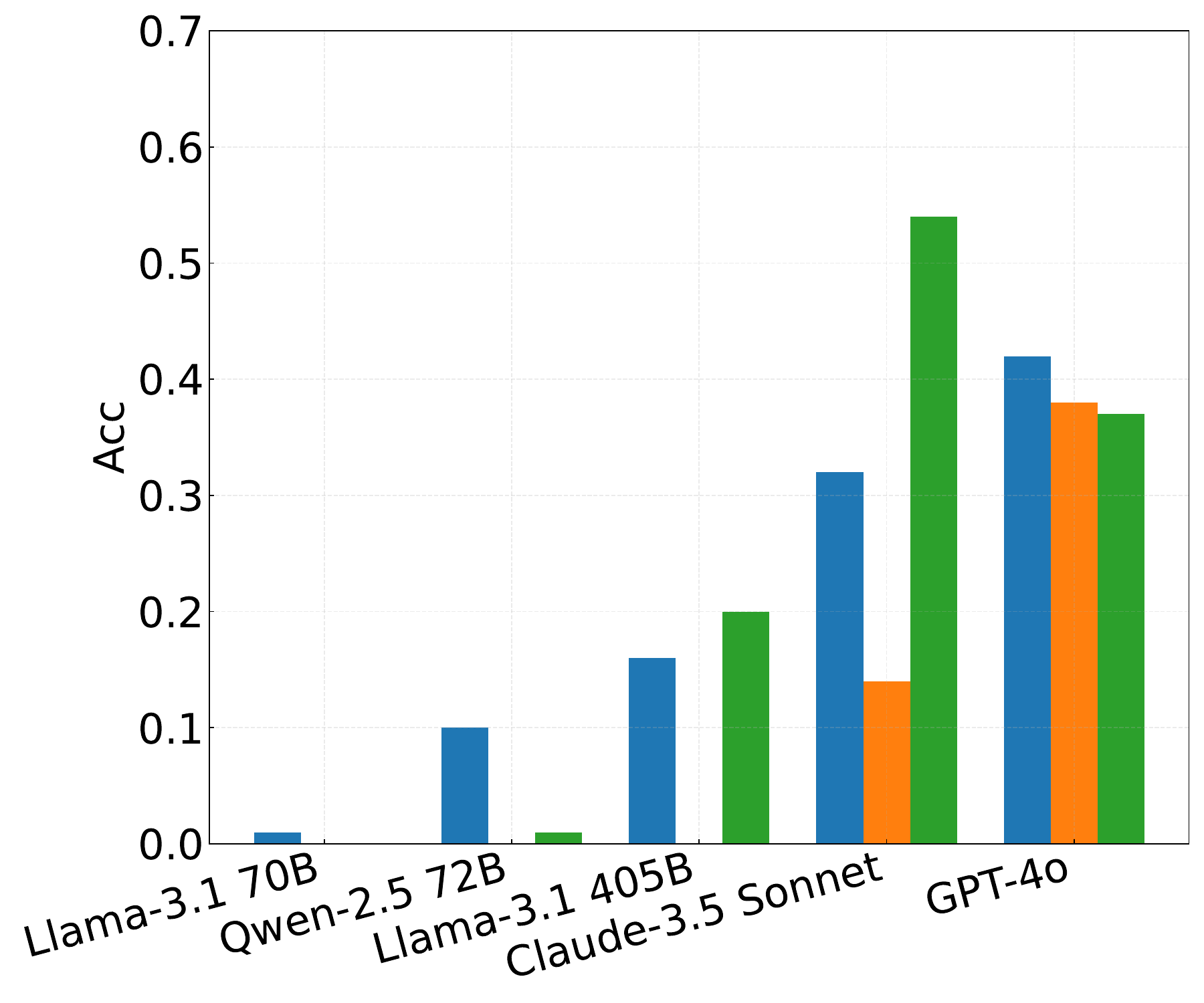}
    \caption{Accuracy}
    \label{fig:context_acc}
  \end{subfigure}
  \caption{The effect of distractive rules and context length. The ``Standard'' mode refers to the default setting of Level 1 tax problems, the ``Distractor'' mode appends nullified forms after the ``Standard'' input, and the ``Placeholder'' mode adds meaningless tokens on space lines. Distractive rules lead to a significant drop on the performances of all LLMs, while meaningless tokens make little difference to the performance.}
  \label{fig:rule_distractor}
\end{figure*}

As shown in Figure \ref{fig:rule_distractor}, the presence of distractive (irrelevant) rules significantly degrades LLM performance, while increasing context length using meaningless placeholders results in only a minor performance drop. These findings suggest that LLMs remain vulnerable to distraction, which undermines their reliability when confronted with superfluous, yet superficially valid, rules.

\subsubsection{Can Tool Augmentation Help?}\label{ap:tool-augmentation}

In Appendix \ref{ap:corr-between-metrics}, we notice that only near-perfect application correctness ${\rm AC}(t)$ can lead to significant accuracy improvement. As the most simple way to reduce errors, especially in mathematical and logical operations, is to introduce external tools, we wonder to what extent external tools can help in our rule-guided reasoning tasks.

Following program of thoughts \citep{chen2023pot} prompting, we ask our LLMs to write Python code to calculate the answer on airline bag fee tasks by defining a \texttt{solution()} function and returning the \texttt{total\_cost} variable, and the execution result of \texttt{solution()} function is viewed as the predicted answer. In this way, the Python interpreter can be viewed as an oracle tool for mathematical and logical calculation. To ensure the correct format of response, we use the 1-shot setting, so we compare the additional results with the original 1-shot results in Table \ref{tab:main_table} as follows:

\begin{table}[!ht]
    \centering
    \renewcommand{\arraystretch}{1.1}
    \resizebox{12cm}{!}{
    \begin{tabular}{lccccccc}
        \toprule
        \multirow{2}{*}{\textbf{Models}} & \multirow{2}{*}{\textbf{Setting}} & \multicolumn{3}{c}{\textbf{Level 1}} & \multicolumn{3}{c}{\textbf{Level 2}} \\
        \cmidrule(lr){3-5} \cmidrule(lr){6-8} & & ${\rm AC}(t)$ & ${\rm R}(t)$ & ${\rm Acc}(t)$ & ${\rm AC}(t)$ & ${\rm R}(t)$ & ${\rm Acc}(t)$\\
        \midrule
        \multirow{2}{*}{Llama 70B} & 1-shot Default & 0.809 & 0.787 & 0.17 & 0.827 & 0.801 & 0.07 \\
         & Tool Augmented & \textbf{0.863} & \textbf{0.882} & \textbf{0.34} & 0.827 & \textbf{0.887} & \textbf{0.18} \\
        \hdashline
        \multirow{2}{*}{Qwen 72B} & 1-shot Default & 0.836 & \textbf{0.908} & 0.19 & 0.818 & \textbf{0.901} & 0.10 \\
         & Tool Augmented & \textbf{0.939} & 0.899 & \textbf{0.42} & \textbf{0.946} & 0.899 & \textbf{0.26} \\
        \hdashline
        \multirow{2}{*}{GPT-4o} & 1-shot Default & 0.922 & 0.885 & 0.32 & 0.875 & 0.853 & 0.16 \\
         & Tool Augmented & \textbf{0.939} & \textbf{0.914} & \textbf{0.44} & \textbf{0.937} & \textbf{0.940} & \textbf{0.33} \\
        \bottomrule
    \end{tabular}
    }
    \caption{Results of different LLMs with tool augmentation on airline tasks.}
    \label{tab:tool_aug}
\end{table}

As can be seen from these results, when provided with oracle math and logic tools, LLMs can achieve a significant performance boost in terms of accuracy ${\rm Acc}(t)$. However, even provided with such tools, LLMs are far from being able to resolve our rule-guided reasoning tasks. We observe non-perfect recall ${\rm R}(t)$ and correctness ${\rm AC}(t)$, which indicates that LLMs still make mistakes in generated codes.

\subsubsection{Summary of Factors that Influence Rule-Guided Following}
In summary, various factors, such as rule complexity, the presence of distractive information, and the difficulty gap between in-context examples and target problems, can profoundly influence LLM performance. Even when LLMs succeed in simpler conditions, challenges like complex mathematical reasoning, large amounts of extraneous rules, and non-ideal in-context samples can severely limit their effectiveness on \textsc{RuleArena} problems.

\section{LLM Prompts}

\paragraph{Airline.} The prompt template we use in airline domain is as follows.

\begin{lstlisting}[style=prompt_style]
(*@\color{codepurple}{\textbf{System Prompt}}@*): You are a helpful assistant at American Airlines.

(*@\color{codepurple}{\textbf{User Prompt}}@*): You are given the information of a passenger, his / her items, his / her special needs, and the policies of American Airlines. You should compute the total cost (including the flight ticket fee, checked bag fees, cost of special needs) according to the policies for the passenger. The policies of American Airlines are as follows:

<reference_rules>

<user_query> Compute the total cost for him step by step (don't omit any bag) and end your response with "The total cost is $xxx." (xxx is a number)
Your response:
\end{lstlisting}

\paragraph{NBA.} The prompt template we use in NBA domain is as follows.

\begin{lstlisting}[style=prompt_style]
(*@\color{codepurple}{\textbf{System Prompt}}@*): You are a helpful NBA team consultant.

(*@\color{codepurple}{\textbf{User Prompt}}@*): You are given rules in NBA Collective Bargaining Agreement and the information about some teams and players. Then you will be given a list of operations, each of which desribes how some teams conduct some transaction. You should determine whether each operation complies with the given rules.

Assume:
* the Salary Cap for the prior (2023-24) Salary Cap Year is $136,000,000;
* the Average Player Salary for the prior (2023-24) Salary Cap Year is $9,700,000;
* the Salary Cap for the current (2024-25) NBA Salary Cap Year is $140,588,000;
* the Luxury Tax is $170,814,000;
* the First Apron Level is $178,132,000;
* the Second Apron Level is $188,931,000;
* the Team Salary of each team listed under "Team Situations:" do not include the amount of contracts that expire at the end of 2023-2024 Salary Cap Year.

Reference Rules in NBA Collective Bargaining Agreement:

<reference_rules>

Decide whether any operation by any team violate the rules:

<user_query>

Analyze the described operations and explicitly state the type of Salary Cap Exceptions if you think the exception should be involved. Conclude your response with:
* "Answer: False." if there is no violation to the rules;
* "Answer: True. Illegal Operation: X. Problematic Team: Y." if Team Y in Operation X violates the rules. Both X and Y should be a single capital letter as A/B/C/...
Your response:
\end{lstlisting}

\paragraph{Tax.} The prompt template we use in tax domain is as follows, where ``<irs\_forms>'' includes both form instructions and user query information.
\begin{lstlisting}[style=prompt_style]
(*@\color{codepurple}{\textbf{System Prompt}}@*): You are a helpful US taxation consultant.

(*@\color{codepurple}{\textbf{User Prompt}}@*): You are given several forms used to report US income tax and the instructions or rules about how to fill the forms. Then you will be given the income and/or payment information about a tax payer According to the given information. You should calculate the income tax owed by this payer.

IRS Forms for the tax payer:

<irs_forms>

Calculate the tax owed by the payer step-by-step according to the information provided by the forms. You should calculate all fields marked with [__]. DO NOT round numbers without explicit instructions. End your response with:
1. "The total tax owed is $xxx." (xxx is a number) if there is tax owed.
2. "The total tax overpaid is $xxx." (xxx is a number) if there is tax overpaid (and should be refunded).
Your response:
\end{lstlisting}